\documentclass[10pt,logo,copyright]{nvidiatechreport}
\usepackage[authoryear,round]{natbib}

\usepackage[utf8]{inputenc} 
\usepackage[T1]{fontenc}    

\usepackage{parskip}        
\usepackage{url}            
\usepackage{booktabs}       
\usepackage{amsfonts}       
\usepackage{nicefrac}       
\usepackage{microtype}      
\usepackage{xcolor}         
\usepackage[dvipsnames]{xcolor} 
\usepackage{graphicx}
\usepackage{animate}        
\usepackage{subcaption}
\usepackage{tabularx}
\usepackage{makecell}
\usepackage{adjustbox}
\usepackage{setspace}
\newcolumntype{M}[1]{>{\centering\arraybackslash}m{#1}}
\usepackage{float}
\usepackage{tikz}
\usetikzlibrary{positioning,shapes,arrows}
\usepackage{amsmath,amsfonts,bm, bbm,leftindex}
\usepackage{multirow}
\usepackage{comment}
\usepackage{gensymb}
\usepackage{lipsum}
\usetikzlibrary{arrows.meta, positioning, fit}
\usepackage[para]{threeparttable}
\usepackage{tikz}
\usetikzlibrary{tikzmark}

\usepackage[most]{tcolorbox}
\usepackage{fancyvrb}
\usepackage{fvextra}
\usepackage{dashrule}

\newcommand{\plus}[1]{\textcolor{ForestGreen}{(+#1)}}

\def\eqref#1{equation~\ref{#1}}

\def\1{\bm{1}}

\DeclareMathAlphabet{\mathsfit}{\encodingdefault}{\sfdefault}{m}{sl}
\SetMathAlphabet{\mathsfit}{bold}{\encodingdefault}{\sfdefault}{bx}{n}

\makeatletter
\let\save@mathaccent\mathaccent
\newcommand*\if@single[3]{%
  \setbox0\hbox{${\mathaccent"0362{#1}}^H$}%
  \setbox2\hbox{${\mathaccent"0362{\kern0pt#1}}^H$}%
  \ifdim\ht0=\ht2 #3\else #2\fi
  }
\newcommand*\rel@kern[1]{\kern#1\dimexpr\macc@kerna}
\newcommand*\widebar[1]{\@ifnextchar^{{\wide@bar{#1}{0}}}{\wide@bar{#1}{1}}}
\newcommand*\wide@bar[2]{\if@single{#1}{\wide@bar@{#1}{#2}{1}}{\wide@bar@{#1}{#2}{2}}}
\newcommand*\wide@bar@[3]{%
  \begingroup
  \def\mathaccent##1##2{%
    \let\mathaccent\save@mathaccent
    \if#32 \let\macc@nucleus\first@char \fi
    \setbox\z@\hbox{$\macc@style{\macc@nucleus}_{}$}%
    \setbox\tw@\hbox{$\macc@style{\macc@nucleus}{}_{}$}%
    \dimen@\wd\tw@
    \advance\dimen@-\wd\z@
    \divide\dimen@ 3
    \@tempdima\wd\tw@
    \advance\@tempdima-\scriptspace
    \divide\@tempdima 10
    \advance\dimen@-\@tempdima
    \ifdim\dimen@>\z@ \dimen@0pt\fi
    \rel@kern{0.6}\kern-\dimen@
    \if#31
      \overline{\rel@kern{-0.6}\kern\dimen@\macc@nucleus\rel@kern{0.4}\kern\dimen@}%
      \advance\dimen@0.4\dimexpr\macc@kerna
      \let\final@kern#2%
      \ifdim\dimen@<\z@ \let\final@kern1\fi
      \if\final@kern1 \kern-\dimen@\fi
    \else
      \overline{\rel@kern{-0.6}\kern\dimen@#1}%
    \fi
  }%
  \macc@depth\@ne
  \let\math@bgroup\@empty \let\math@egroup\macc@set@skewchar
  \mathsurround\z@ \frozen@everymath{\mathgroup\macc@group\relax}%
  \macc@set@skewchar\relax
  \let\mathaccentV\macc@nested@a
  \if#31
    \macc@nested@a\relax111{#1}%
  \else
    \def\gobble@till@marker##1\endmarker{}%
    \futurelet\first@char\gobble@till@marker#1\endmarker
    \ifcat\noexpand\first@char A\else
      \def\first@char{}%
    \fi
    \macc@nested@a\relax111{\first@char}%
  \fi
  \endgroup
}
\makeatother

\usepackage{pifont}

\usepackage[nameinlink]{cleveref}
\crefname{equation}{Eq.}{Eqs.}
\crefname{figure}{Fig.}{Figs.}
\crefname{section}{Sec.}{Sec.}
\crefname{appendix}{App.}{App.}
\crefname{table}{Tab.}{Tabs.}
\crefname{algorithm}{Algo}{Algo}
\crefname{thm}{Thm}{Thm}
\Crefname{thm}{Thm}{Thm}
\crefname{prop}{Prop}{Prop}

\definecolor{darkred}{rgb}{0.7, 0.0, 0.0}

\newcommand{\cosmosreason}{Cosmos-Reason1\xspace}

\newcommand{\crefnames}[3]{%
  \@for\next:=#1\do{%
    \expandafter\crefname\expandafter{\next}{#2}{#3}%
  }%
}

\title{\cosmosreason: From Physical Common Sense To Embodied Reasoning}

\author{NVIDIA\footnote{A detailed list of contributors and acknowledgments can be found in~\cref{sec::contributors} of this paper.}}

\begin{abstract}
Physical AI systems need to perceive, understand, and perform complex actions in the physical world. In this paper, we present the \cosmosreason models that can understand the physical world and generate appropriate embodied decisions (\eg, next step action) in natural language through long chain-of-thought reasoning processes. We begin by defining key capabilities for Physical AI reasoning, with a focus on physical common sense and embodied reasoning. To represent physical common sense, we use a hierarchical ontology that captures fundamental knowledge about space, time, and physics. For embodied reasoning, we rely on a two-dimensional ontology that generalizes across different physical embodiments.
Building on these capabilities, we develop two multimodal large language models, \cosmosreason-7B and \cosmosreason-56B.
We curate data and train our models in two stages: Physical AI supervised fine-tuning (SFT) and Physical AI reinforcement learning (RL).
To evaluate our models, we build comprehensive benchmarks for physical common sense and embodied reasoning according to our ontologies. Evaluation results show that Physical AI SFT and RL bring significant improvements. To facilitate the development of Physical AI, we make our code and pre-trained models available under the NVIDIA Open Model License at \url{https://github.com/nvidia-cosmos/cosmos-reason1}.
\end{abstract}

\begin{document}

\maketitle

\abscontent

\section{Introduction}
\label{sec::intro}

Physical AI systems are designed to interact with the physical world. To effectively follow instructions and take appropriate actions to achieve a desired goal, they need first to perceive, understand, and reason about the physical world. Recently, with breakthroughs of post-training and test-time scaling via long chain-of-thought (CoT) processes, large language models (LLMs) have demonstrated remarkable general reasoning capabilities in tackling complex problems across domains such as coding and mathematics~\citep{openaio1,deepseekai2025deepseekr1}. However, a key limitation of these models lies in their ability to ground their knowledge in the physical world. While LLMs trained on vast amounts of textual data from the Internet may acquire essential knowledge to reason about the physical world, they often struggle to establish connections between that knowledge and real-world interactions and dynamics.

In this paper, we begin by defining the fundamental capabilities essential for Physical AI systems. Unlike designing models that excel at solving coding and mathematical problems, our focus is on empowering models with physical common sense knowledge and embodied reasoning capabilities grounded in the real world. To establish a shared framework and help measure progress, we propose two ontologies. First, a hierarchical ontology that organizes physical common sense into three major categories --- Space, Time, and Fundamental Physics --- which are further divided into 16 fine-grained subcategories. This ontology encapsulates knowledge about how the physical world operates under the laws of physics and how it responds to interactions with embodied agents. Second, we introduce a two-dimensional ontology for embodied reasoning, which encompasses four key reasoning capabilities across five types of embodied agents. Effective embodied reasoning based on physical common sense is crucial for understanding and planning actions that achieve specific objectives in the real world. The details are described in \Cref{sec::definition}.

We introduce \textit{\cosmosreason} as a step toward enabling multimodal LLMs to generate more physically grounded responses. We focus on the visual world, where the observations of the world are represented as videos. \cosmosreason perceives the physical world through video input, understands it, and reasons about it through long chain-of-thought thinking processes before generating responses. These responses, expressed in natural language, include both explanatory insights and embodied decisions, such as determining the next action to take. We employ a decoder-only multimodal LLM architecture where the input video is processed by a vision encoder followed by a projector to align with the text token embeddings before feeding into the LLM. 
We experiment with both dense Transformer and hybrid Mamba-MLP-Transformer architecture as the LLM backbone.
\cosmosreason comes with two model sizes: \cosmosreason-7B and \cosmosreason-56B.
We describe details about the model architecture in~\cref{sec::model}.

Constructing rule-based, verifiable rewards at scale has been critical for the reasoning LLMs' success in solving math and coding problems. Can we design rule-based, verifiable rewards for training Physical AI reasoning models with reinforcement learning? In this work, we explore two types of rewards based on answering multiple-choice questions (MCQs). The first type of MCQs are designed based on human annotations. With inspiration from video self-supervised learning, we automatically generate a second type of MCQ based on the structure of video data itself, such as solving puzzles with shuffled spatiotemporal video patches or predicting the arrow of time on whether a video is playing forward or backward. All these rewards are rule-based, verifiable, and highly relevant to Physical AI capabilities. We discuss details about the design of RL training algorithm and infrastructure in~\cref{sec::rl}.

Data determines our model's ceiling. 
To strengthen our models' physical common sense and embodied reasoning capabilities, we carefully curate around 4M annotations of video and text pairs, including captions for physical understanding, multiple choice questions, and long chain-of-thought reasoning traces.
We develop two pipelines to curate these physical common sense and embodied reasoning data according to our ontologies. The data are curated based on human annotations and model distillation from DeepSeek-R1~\citep{deepseekai2025deepseekr1} for Physical AI supervised fine-tuning.
Details about data are discussed in~\cref{sec::data}.

\begin{figure}[!t]
    \centering
    \includegraphics[width=0.95\textwidth]{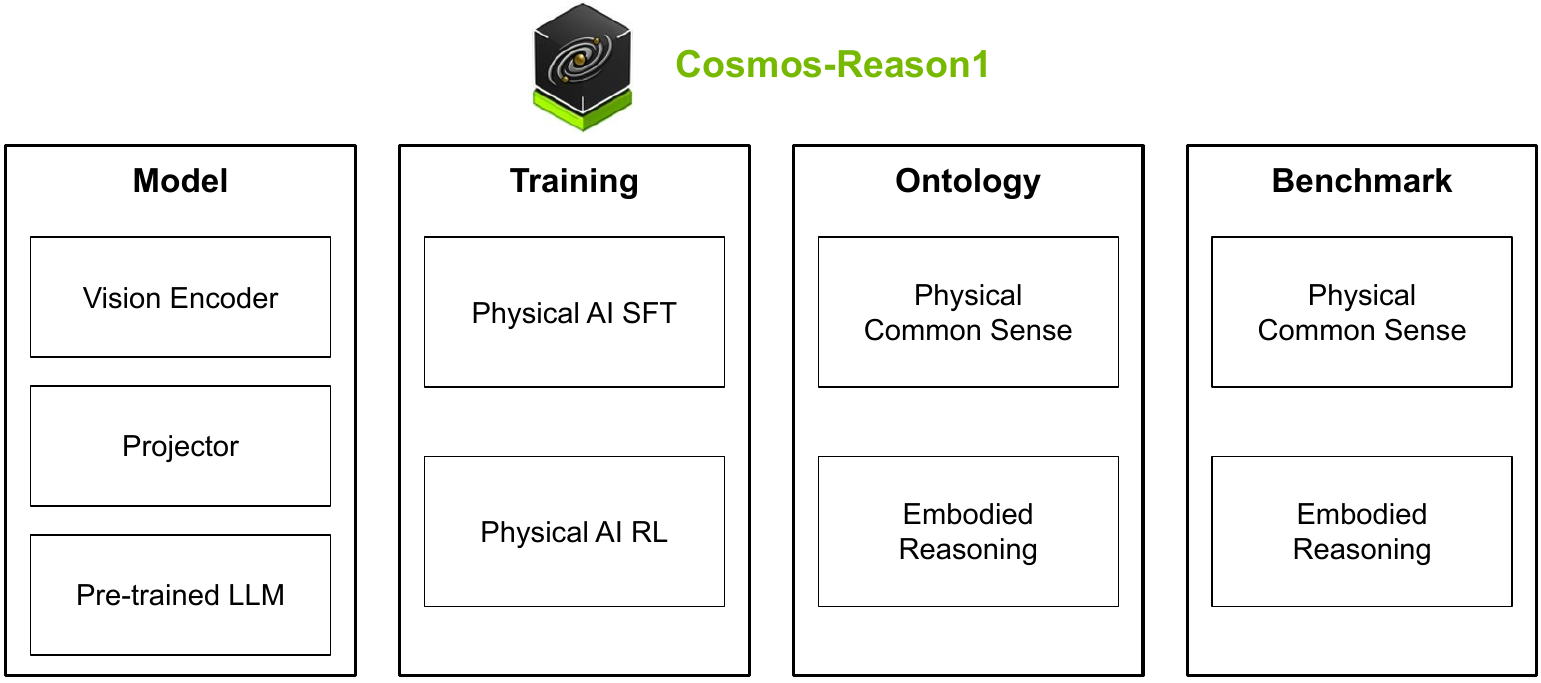}
    \caption{An overview of \cosmosreason. \cosmosreason contains two multimodal large language models of 7B and 56B, trained in two stages, including Physical AI SFT and Physical AI RL. We also define two ontologies for physical common sense and embodied reasoning, and build two benchmarks to evaluate models' Physical AI reasoning capabilities.}
    \label{fig:teaser}
    \vspace{-2mm}
\end{figure}

In order to evaluate our models, we build new benchmarks for evaluating Physical AI capabilities in~\cref{sec::benchmark}. For physical common sense in~\cref{sec::benchmark_physical_reasoning}, we build three benchmarks (Space, Time, and Fundamental Physics) containing 604 questions from 426 videos. For embodied reasoning in~\cref{sec::benchmark_embodied_reasoning}, we build six benchmarks containing 610 questions from 600 videos, covering a wide range of tasks across different physical embodiments, including humans, robot arms, humanoid robots, and autonomous vehicles. 

\cref{sec::experiment} presents the evaluation results of \cosmosreason and comparisons with existing models. In~\cref{sec::physical_ai_sft}, we introduce the experiment setup, including training details for Physical AI SFT, and evaluation results of SFT models on our benchmarks. In~\cref{sec::exp_rl}, we present the evaluation results for Physical AI RL. Training using our rule-based, verifiable rewards in RL post-training leads to improvements in all of our benchmarks. 

\cref{fig:teaser} shows an overview of \cosmosreason. 
In summary, we introduce two multimodal large language models, \cosmosreason-7B and \cosmosreason-56B.
The models are trained in two stages: Physical AI SFT and Physical AI RL. We define ontologies for physical common sense and embodied reasoning and build benchmarks to evaluate models' Physical AI reasoning capabilities. To facilitate the advancement of Physical AI, we make our code and pre-trained models available under the NVIDIA Open Model License at \url{https://github.com/nvidia-cosmos/cosmos-reason1}. Building reasoning models for Physical AI is an open problem that is far from being solved, and we hope our paper contributes to the advancement of this field.

\section{Physical AI Reasoning}
\label{sec::definition}

We identify two important capabilities for Physical AI reasoning models --- physical common sense reasoning and embodied reasoning. 
First, Physical AI models should possess physical common sense, meaning a general, embodiment-agnostic understanding of the environment and forms the basis for predicting what is plausible and implausible in the real world. Second, Physical AI models should also help embodied agents perceive, reason, and make decisions about planning future interactions with the physical environment. We seek to incorporate both ``System 1'' and ``System 2'' in physical common sense reasoning and embodied reasoning. ``System 1'' enables fast, intuitive responses such as pattern recognition and instinctive judgments, while ``System 2'' operates more slowly, engaging in deliberate reasoning for complex decision-making~\citep{kahneman2011thinking}.
\subsection{Common Sense Reasoning}
\label{sec::definition_common_sense}

Humans acquire physical common sense primarily through passive observation of the world. For example, infants can understand basic concepts such as object permanence and gravity in a few months after birth~\citep{riochet2021intphys}.
This common sense encompasses a collection of knowledge about what is possible, impossible, or likely to happen in the real world.
Training AI systems in real-world environments is expensive and can pose risks to both the system and its surroundings.
Utilizing physical common sense, AI systems can quickly learn new skills with minimal trial and error while avoiding making critical mistakes in uncertain scenarios~\citep{lecun2022path}.

To define physical common sense, we introduce an ontology comprising three broad categories: Space, Time, and other Fundamental Physics, further divided into 16 fine-grained subcategories. Inspired by~\citet{morrisposition}, we focus on capabilities rather than processes. 
Specifically, our ontology identifies key capabilities that Physical AI models should possess, without specifying the mechanisms or embodiments by which a system accomplishes tasks.
For example, we believe the ability to understand the spatial relationship of objects, the temporal order of events, and object permanence are fundamental to Physical AI. However, such systems need not necessarily act human-like, such as grasping with dexterous hands with fingers or walking on two legs.

\begin{figure}[!htp]
    \centering
    \includegraphics[width=0.5\textwidth]{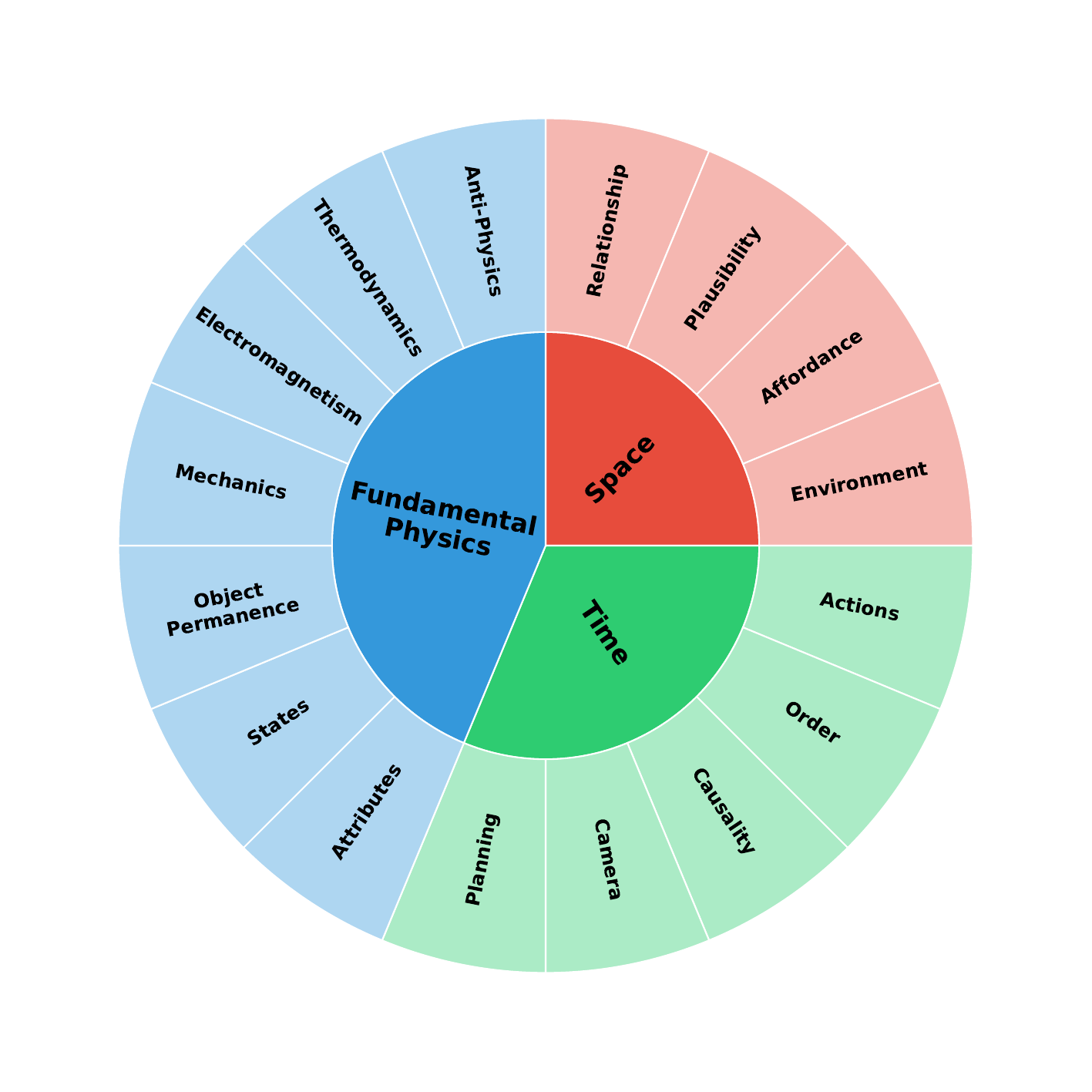}
    \caption{A pie chart showing our physical common sense ontology. The ontology has three categories (Space, Time, and Fundamental Physics) and 16 fine-grained subcategories.}
    \label{fig:ontology}
\end{figure}

We show our physical common sense ontology in~\cref{fig:ontology}. The Space category encompasses the relationships between objects, their interactions, and the surrounding environment. 
It includes concepts such as Relationship, Plausibility, Accordance, and Environment.
The Time category pertains to actions and events that unfold over a duration, covering Actions, Order, Causality, Camera, and Planning.
Lastly, we introduce a Fundamental Physics category to address objects and core physical principles, including Attributes, States, Object Permanence, Mechanics, Electromagnetism, Thermodynamics, and Anti-Physics. Detailed definitions of all subcategories are described in~\cref{tab:common_sense_ontology}.

\begin{table}[ht]
    \small
    \captionsetup{justification=centering}
    \caption{Definition of Physical AI systems' capabilities for each subcategory in our common sense ontology.}
    \centering
    \begin{tabular}{p{3.5cm}p{12.25cm}}
        \toprule
        \textbf{Category: Subcategory} & \textbf{Capability}  \\
        \midrule
        \textbf{Space: Relationship} & Determine the spatial relationship of objects in a scene. Perspective is important; for example, an object is left to the person or left to the camera view. \\
        \midrule
        \textbf{Space: Plausibility} & Determine if a possible spatial relationship is feasible. \\
        \midrule
        \textbf{Space: Affordance} & Understand object interaction with subjects such as humans, animals, robots, \etc. \\
        \midrule
        \textbf{Space: Environment} & Understand the scene or the surrounding environment. \\
        \midrule
        \textbf{Time: Actions} & Understand actions, including the ability to accurately describe the action (movement, direction, intensity, \etc.), determine action objective, subtask or goal decomposition, and determine if a task/objective is successfully completed. \\
        \midrule
        \textbf{Time: Order} & Understand the timestamp and sequential order of events. \\
        \midrule
        \textbf{Time: Causality} & Understand if event A causes B. \\
        \midrule
        \textbf{Time: Camera} & Determine the position and movement of the camera, including camera movement, camera angle/position, and transition of scenes. \\
        \midrule
        \textbf{Time: Planning} & Come up with a future plan based on past observations. \\
        \midrule
        \textbf{Fundamental Physics: \newline Attributes} & Determine physical properties of an object, including semantic description, size, color, material, mass, temperature, solidity (can objects pass through one another?), \etc. \\
        \midrule
        \textbf{Fundamental Physics: \newline States} & Determine the object state and understand the state change (\eg, ice changed to water, eggs changed from raw to cooked). \\
        \midrule
        \textbf{Fundamental Physics: \newline Object Permanence} & Understand object permanence, which properties can/cannot change in certain conditions (weight, shape, size, color, \etc.). \\
        \midrule
        \textbf{Fundamental Physics: \newline Mechanics} & Understand laws of physics related to Mechanics, including Statics (balance, stability, support, elasticity, deformation, the center of mass, \etc), Kinematics (velocity, acceleration, linear motion, circular motion, rotational motion, \etc), and Dynamics (gravity, collision, friction, sliding, inertia, conservation of momentum, fluids and particles, \etc). \\
        \midrule
        \textbf{Fundamental Physics: \newline Electromagnetism} & Understand laws of physics related to Electromagnetism, including Optics (lighting, shadow, occlusion, reflection, refraction, diffraction, absorption, transmission, \etc), Electricity, and Magnetism. \\
        \midrule
        \textbf{Fundamental Physics: \newline Thermodynamics} & Understand laws of physics related to Thermodynamics, such as heat, temperature change, evaporation, heat transfer, thermal expansion and contraction, \etc. \\
        \midrule
        \textbf{Fundamental Physics: \newline Anti-Physics} & Understand situations that defy the laws of physics, such as anti-gravity, reverse of time, perpetual motion, sudden disappearance, \etc. \\
        \bottomrule
    \end{tabular}
    \label{tab:common_sense_ontology}
\end{table}

\subsection{Embodied Reasoning}
\label{sec::definition_embodied_reasoning}

Physical AI operates in the real world, where environments are dynamic, uncertain, and governed by complex physical interactions. Unlike \textit{abstract reasoning} in mathematics and programming, which manipulates symbols in a structured and deterministic manner, embodied reasoning requires an AI system to interact with and learn from the physical world. Unlike \textit{passive understanding}, reasoning in embodied AI is often grounded in action, enabling robots not only to comprehend what they currently observe but also plan intelligent behaviors for the future in uncertain and dynamic environments. Specifically, embodied reasoning requires the capability to: 
\begin{enumerate}
\item \textbf{Process Complex Sensory Inputs.}
Unlike symbolic reasoning, which works with clean data representations, embodied reasoning must extract meaningful patterns from raw, often incomplete, and ambiguous sensory inputs.
\item \textbf{Predict Action Effects.}
Actions have physical consequences, and effective reasoning requires an intuitive grasp of cause-and-effect relationships. An AI system must predict how an object will respond to force, how a robot’s body will interact with its surroundings, or how a vehicle’s movement will be influenced by terrain and physics.
\item \textbf{Respect Physical Constraints.}
Unlike abstract problem-solving, which often involves optimizing discrete choices, embodied reasoning must account for real-world physics, such as inertia, friction, and material properties. It requires AI to generate long-horizon action plans that are feasible given physical constraints, ensuring stability, efficiency, and safety in execution.
\item \textbf{Learn from Interaction.}
In Physical AI, actions do not occur in isolation; every movement or decision affects the environment and generates feedback. Embodied reasoning must continuously update its understanding based on these interactions, allowing the system to refine its behavior dynamically.
\end{enumerate}

Embodied reasoning is also not confined to a single type of agent — it is applicable to humans, animals, and robots across various forms (\eg, robotic arms, humanoid figures, or autonomous vehicles). They all need to develop similar embodied reasoning skills to navigate, manipulate, and make adaptive decisions under different environmental conditions and task goals. 
We summarize capabilities and types of physical embodiments with examples into a two-dimensional ontology in~\cref{tab:embodied_reasoning_ontology}.

In this paper, we focus on the first three embodied reasoning capabilities we defined above and leave ``Learn from Interactions'' as future work.
Specifically, we focus on video input as a representative example of ``Process Complex Sensory Inputs''.
For ``Predict Action Effects'', we focus on two tasks, including task-completion verification for determining whether a task has been completed, and the next plausible action prediction for predicting the next plausible next action to achieve a goal.
For ``Respect Physical Constraints'', we focus on action affordance to assess whether it is possible to perform a specific action toward a goal.
We collect videos across different agents, including humans, robot arms, humanoid robots, and autonomous vehicles. By investigating these varied cases, we aim to deepen our understanding of how embodied reasoning enables intelligent interaction with the physical world.

\begin{table}[ht]
    \small
    \captionsetup{justification=centering}
    \caption{Embodied reasoning ontology, with an example for each combination of capability and agent type.}
    \centering
    \begin{tabular}{p{2.5cm}|p{5.25cm}p{7.5cm}}
        \toprule
         & \textbf{Natural Agents \newline (humans, animals)} & \textbf{Robotics Systems \newline (robot arms, humanoid robots, autonomous vehicles)} \\
        \midrule
        \textbf{Process Complex \newline Sensory Inputs} & A person watches videos about a cooking recipe. \newline A bat locates prey using echolocation. & A robot arm recognizes objects using its camera. \newline A robot detects obstacles while walking. \newline A self-driving car recognizes a stop sign and pedestrians. \\
        \midrule
        \textbf{Predict Action \newline Effects} & A carpenter anticipates wood splintering before cutting. \newline A dog estimates a ball’s landing spot to catch it. & A robotic arm compensates for momentum before gripping an object. \newline A robot estimates an object’s weight before lifting it. \newline A self-driving car predicts tire slippage on ice. \\
        \midrule
        \textbf{Respect Physical \newline Constraints} & A pilot maintains altitude within aerodynamic limits. \newline A cheetah limits speed to avoid muscle strain. & A robotic gripper limits its force to prevent breaking objects. \newline A robot adjusts joint torque to prevent falls. \newline A drone avoids exceeding wind resistance thresholds. \\
        \midrule
        \textbf{Learn from \newline Interactions} & A golfer corrects their stance after observing ball trajectory. \newline A dog learns to open doors through repeated attempts. & A factory robot improves alignment after detecting misplacements. \newline A robot learns new handshakes. \newline A self-driving car refines braking distances. \\
        \bottomrule
    \end{tabular}
    \label{tab:embodied_reasoning_ontology}
\end{table}

\section{\cosmosreason}
\label{sec::model}

\cosmosreason is a family of multimodal large language models specialized in Physical AI reasoning. The family comprises two models: \cosmosreason-7B and \cosmosreason-56B. In this section, we introduce the design of our multimodal architectures and the choices of LLM backbones.

\subsection{Multimodal Architecture}
\label{sec::vlm_arch}

There are different architecture designs for building multimodal large language models~(LLMs) using existing text-only LLM backbones and vision encoders.
Commonly used architectures are the decoder-only architecture (\eg, LLaVA~\citep{liu2023visual}) and the cross-attention-based architecture (\eg, Flamingo~\citep{alayrac2022flamingo} and Llama 3-V~\citep{grattafiori2024llama}).
We utilize the decoder-only architecture similar to LLaVA~\citep{liu2023visual} and NVLM-D~\citep{dai2024nvlm} for its simplicity and unified handling of all modalities by aligning other modality tokens~(image or video) into the text token embedding space.
Specifically, the model begins with the vision encoder~\citep{chen2024internvl}, followed by the projector containing a downsampling two-layer MLP, and then the decoder-only LLM backbone~\citep{adler2024nemotron, waleffe2024empirical, deepseekai2025deepseekr1}.

\begin{figure}[!htp]
    \centering
    \includegraphics[width=0.95\textwidth]{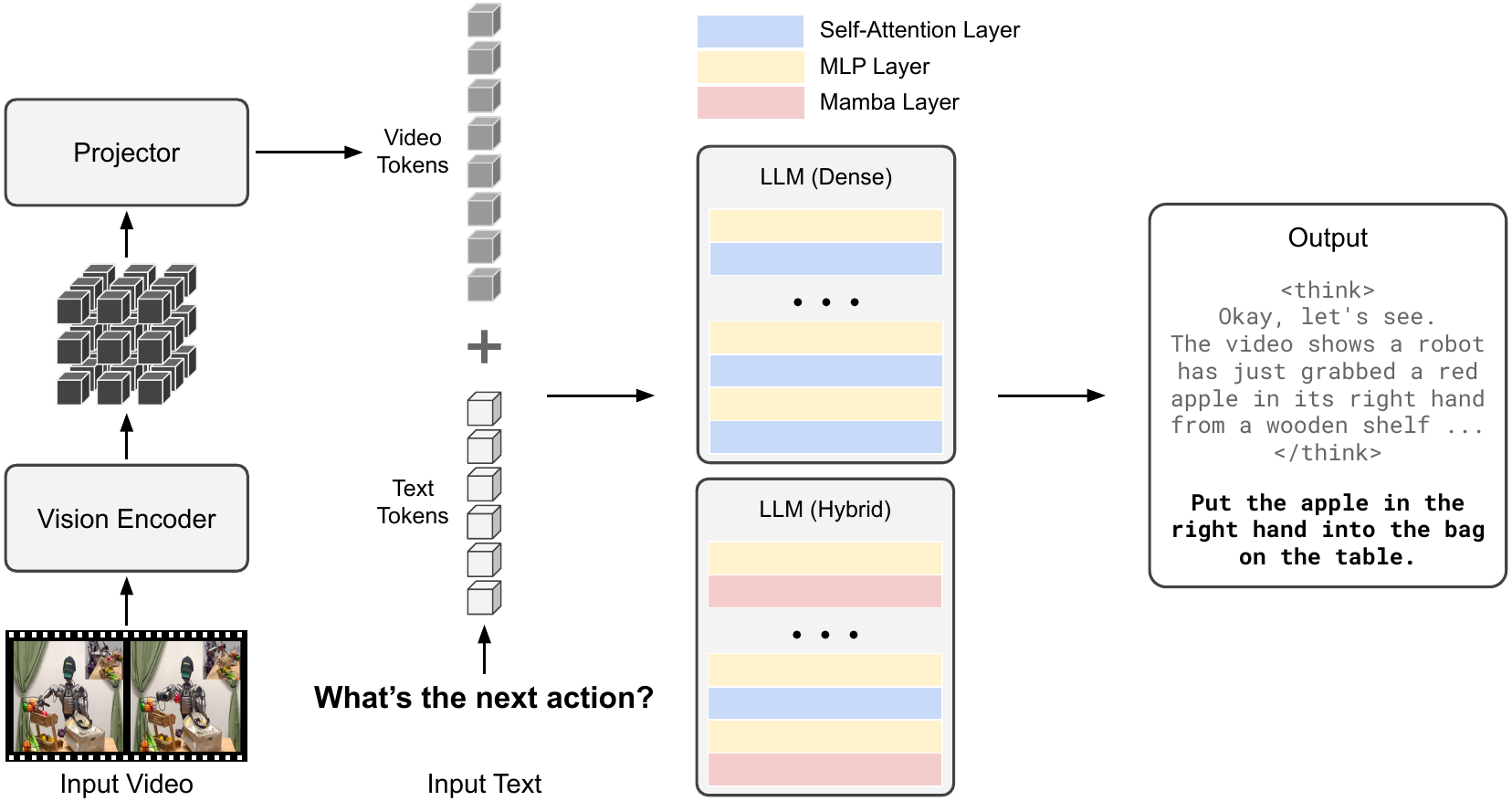}
    \caption{An illustration of our multimodal large language model architecture. Given an input video and an input text prompt, the video is projected into the LLM's token embedding space as video tokens by a vision encoder followed by a projector. The text tokens are concatenated with the video tokens and fed into the LLM backbone, a dense Transformer or a hybrid Mamba-MLP-Transformer architecture. Our model can output responses with long chain-of-thought reasoning processes.}
    \label{fig:vlm_architecture}
\end{figure}

For \cosmosreason-7B, we choose Qwen2.5-VL~\citep{bai2025qwen2} as our pre-trained model and follow the same image and video processing.
For \cosmosreason-56B, we leverage InternViT-300M-V2.5~\citep{chen2024expanding} as our vision encoder and Nemotron-H~\citep{nvidia2025nemotronhfamilyaccurateefficient} as our LLM backbone. 
We perform the following processing for the \cosmosreason-56B hybrid model.
For each input image, we dynamically adjust the image to a predefined aspect ratio and segment it into 1 to 12 tiles, each measuring $448\times448$ pixels, depending on the image's resolution. Additionally, we generate a thumbnail tile; a scaled-down version of the full image to preserve the global context. More details can be found in~\citet{dai2024nvlm}.
For each input video, we uniformly sample up to 32 frames at a maximum rate of 2 frames per second, resizing each frame to $448\times448$ pixels. For each $448\times448$ video frame input, the vision encoder generates 1,024 visual tokens with the patch size of $14\times14$, which are then downsampled by a factor of $2\times2$ using PixelShuffle~\citep{shi2016real}, reducing them to 256 tokens by transforming spatial dimensions into channel dimensions.
The image tokens from multiple tiles are concatenated with interleaved tile ID tags, as described in~\citet{dai2024nvlm}, while the video tokens from multiple frames are concatenated directly.
More discussions about the hybrid LLM backbone are in~\cref{sec::hybrid_model}.
We illustrate our hybrid multimodal architecture in~\cref{fig:vlm_architecture} and summarize our model configurations in~\cref{tab:model_summary}.

\begin{table}[!t]
    \small
    \captionsetup{justification=centering}
    \caption{
        Configuration details of \cosmosreason models.
    }
    \centering
    \setlength{\tabcolsep}{6pt}
    \begin{tabular}{l|cc}
    \toprule
        \textbf{Configuration} & \textbf{\cosmosreason-7B} & \textbf{\cosmosreason-56B} \\
        \midrule
        & \multicolumn{2}{c}{Vision Encoder} \\
        \midrule
        Architecture & ViT-676M & ViT-300M \\
        Input Size & Dynamic & $448\times448$ \\
        Patch Size & $14\times14$ & $14\times14$ \\
        Number of Layers & $32$ & $24$ \\
        Model Dimension & $1{,}280$ & $1{,}024$ \\
        FFN Hidden Dimension & $3{,}456$ & $4{,}096$ \\
        \midrule
        & \multicolumn{2}{c}{Projector} \\
        \midrule
        Downsampling (HxWxT) & $2\times2\times2$ & $2\times2\times1$ \\
        Number of Layers & $2$ & $2$ \\
        Input Dimension & $1{,}280$ & $4{,}096$ \\
        Hidden Dimension & $5{,}120$ & $32{,}768$ \\
        Output Dimension & $3{,}584$ & $8{,}192$ \\
        \midrule
        & \multicolumn{2}{c}{LLM Backbone} \\
        \midrule
        Architecture & Transformer & Mamba-MLP-Transformer \\
        Number of Layers & $28$ & $118$ \\
        Model Dimension & $3{,}584$ & $8{,}192$ \\
        FFN Hidden Dimension & $18{,}944$ & $32{,}768$ \\
        Number of Attention Heads & $28$ & $64$ \\
    \bottomrule
    \end{tabular}
    \label{tab:model_summary}
\end{table}

\subsection{Hybrid Mamba-MLP-Transformer Backbone}
\label{sec::hybrid_model}

Since its introduction, the Transformer architecture~\citep{vaswani2017attention} has revolutionized the field of language modeling, becoming the de facto standard for building foundation models. However, its self-attention mechanism has a quadratic time complexity with respect to its context length. In contrast, the recently proposed Mamba architecture~\citep{gu2023mamba} introduces linear-time sequence modeling with selective state space models, making it significantly more efficient for handling long sequences.
In practice, the selective state spaces of Mamba may not be sufficient to capture every detail within long sequences. To address this, a small portion of Transformer layers is incorporated for long-context modeling, giving rise to the hybrid Mamba-MLP-Transformer architecture~\citep{waleffe2024empirical}.

\begin{figure}[!t]
    \centering
    \includegraphics[width=0.95\textwidth]{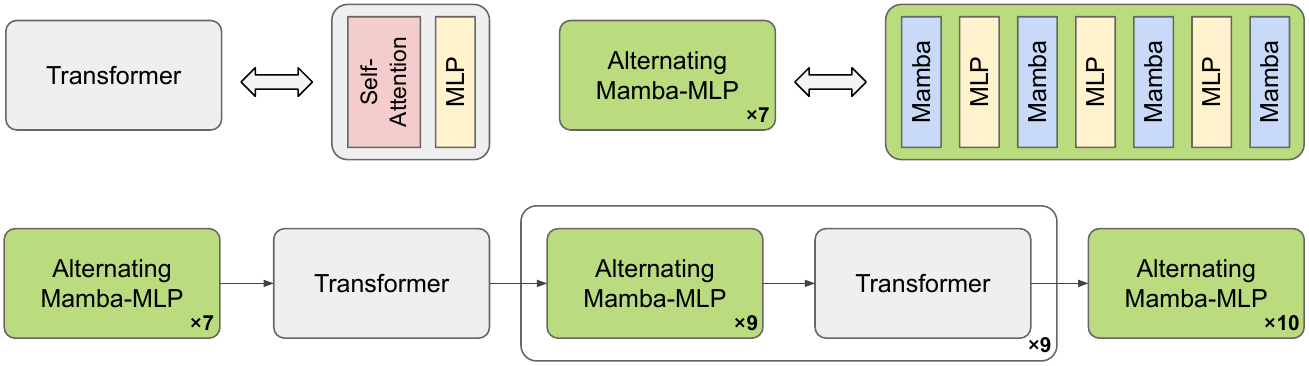}
    \caption{An illustration of our hybrid Mamba-MLP-Transformer backbone architecture used in \cosmosreason-56B. A Transformer block consists of a self-attention layer and an MLP layer. We also show an example of an Alternating Mamba-MLP module on top of the figure.}
    \label{fig:hybrid_model}
\end{figure}

In \cosmosreason-56B, we use a pre-trained model with a hybrid Mamba-MLP-Transformer architecture~\citep{nvidia2025nemotronhfamilyaccurateefficient} as our backbone.
An illustration of the 56B LLM architectures can be found in~\cref{fig:hybrid_model}.
We train the \cosmosreason-7B dense model with a Tensor Parallelism of 4 (TP=4)~\citep{shoeybi2019megatron}, while the \cosmosreason-56B hybrid model is trained with a Tensor Parallelism of 8 and a Pipeline Parallelism of 2 (TP=8, PP=2).

\section{Reinforcement Learning}
\label{sec::rl}

We employ two training stages, Physical AI SFT and Physical AI RL, to adapt a pre-trained visual language model (\eg, Qwen2.5-VL~\citep{bai2025qwen2} or Nemotron-H-VLM~\citep{nvidia2025nemotronhfamilyaccurateefficient}) to a Physical
AI reasoning model. After the SFT stage, we post-train our models using RL with
Physical AI focused tasks to further enhance their physics common sense and embodied reasoning abilities. In
this section, we explain the algorithm and the customized training framework, which highly efficient and robust for RL training.
\begin{figure}[!htp]
    \centering
    \includegraphics[width=0.75\textwidth]{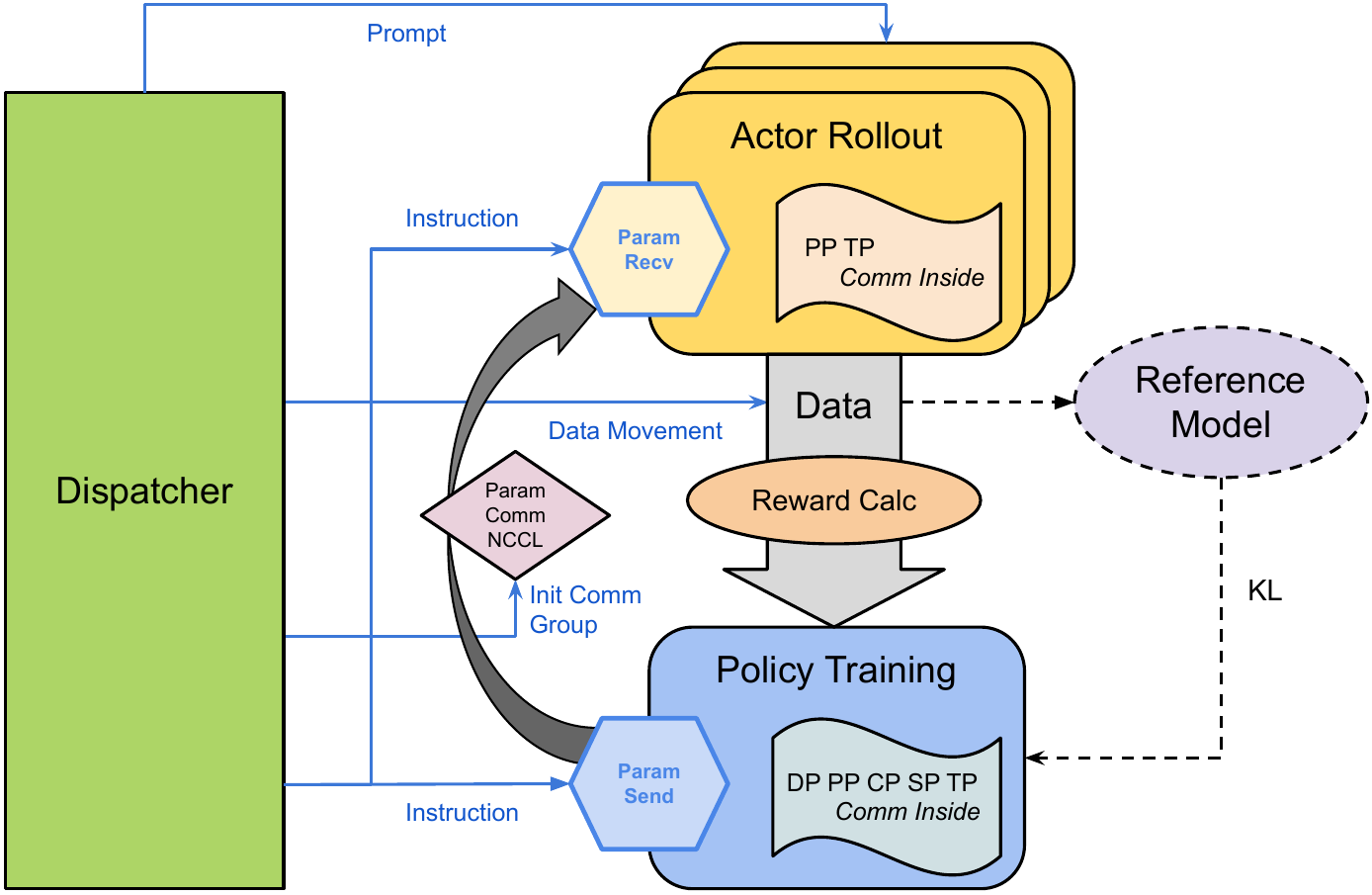}
    \caption{The architecture of the proposed RL training framework. There are mainly three parts of the framework. (1) Dispatcher: schedule and distribute training data, manage the status of the framework. (2) Actor Rollout: generate responses from the prompt, compute rewards and advantages for the policy training. (3) Policy Training: conduct an RL algorithm on the actor and optimize the actor based on the rewards and other constraints. The policy training nodes support 5D-parallelism, \emph{i.e.} data parallel (DP), pipeline parallel (PP), context parallel (CP), fully-sharded data parallel (FSDP), and tensor parallel (TP). The actor rollout nodes support DP, PP, and TP. We adopt a customized NCCL communicator to communicate between the dispatcher and the rollout/policy node.}
    \label{fig:rl_infra}
\end{figure}

\subsection{Algorithm}
We adopt GRPO~\citep{shao2024deepseekmathpushinglimitsmathematical} as our RL algorithm of choice for its simplicity and computational efficiency, as it circumvents the necessity of training and maintaining a separate critic model. GRPO enables a streamlined approach to policy optimization, wherein the advantage function is derived by normalizing rewards within a cohort of responses generated for each prompt. Let $R(o_i)$ denotes the reward for response $o_i$, in a group of responses $\mathcal{G} = \{o_1, o_2, \dots, o_G\}$, then the computed advantage can be expressed as:
$$
A_i = \frac{R(o_i) - \text{mean}(\mathcal{G})}{\text{std}(\mathcal{G})}
$$

\subsection{Training Framework}
To make more efficient use of RL training data, we also propose a novel, fully asynchronous and highly robust RL training framework. The architecture is shown in \cref{fig:rl_infra}. First, unlike mainstream colocated frameworks \citep{hu2024openrlhf, sheng2024hybridflow} that suffer from inefficient resource utilization due to synchronization overhead, the proposed framework adopts a heterogeneous deployment strategy for policy training and actor rollout. By leveraging a unified dispatcher to schedule and dispatch training prompts, our design enables asynchronous parallelism across the entire training pipeline. This achieves end-to-end asynchrony while maintaining training effectiveness, resulting in approximately a 160\% improvement in training efficiency compared with the colocated frameworks. Second, attributed to our novelly designed training mesh management logic, the framework is able to rapidly reconfigure itself and continue the current training step even if any node fails during training, without requiring costly restarts or recovery procedures. Furthermore, the dispatcher redundancy mechanism also improves overall framework robustness, and the fault-tolerant design inherently supports dynamic scaling, both up and down, enabling flexible resource allocation based on workload demands.
\section{Data}
\label{sec::data}

In this section, we explain the data sources and their curation procedure of Physical AI SFT and RL for training Cosmos-Reason1 models.

\subsection{Physical AI Supervised Fine-Tuning}
\label{sec::physicalai_sft}

In this stage, we fine-tune models from the previous phase on domain-specific data to specialize in Physical AI. This process aims to achieve two key outcomes: (1) enhance the model's vision-language capabilities on Physical AI-specific data and (2) develop two critical reasoning abilities --- physical common sense reasoning and embodied reasoning (detailed in \Cref{sec::definition_common_sense} and \Cref{sec::definition_embodied_reasoning}). Unlike the previous two training stages, existing data sources cannot be directly utilized for Physical AI SFT. To address this challenge, we develop a specialized pipeline to carefully curate SFT datasets for both physical common sense and embodied reasoning applications.
A portion of the Physical AI SFT data, particularly visual question-answering (VQA) pairs, are generated through a model-in-the-loop approach rather than directly from human curation.

\begin{figure}[!htp]
    \centering
    \subfloat{
        \includegraphics[height=3.36cm]{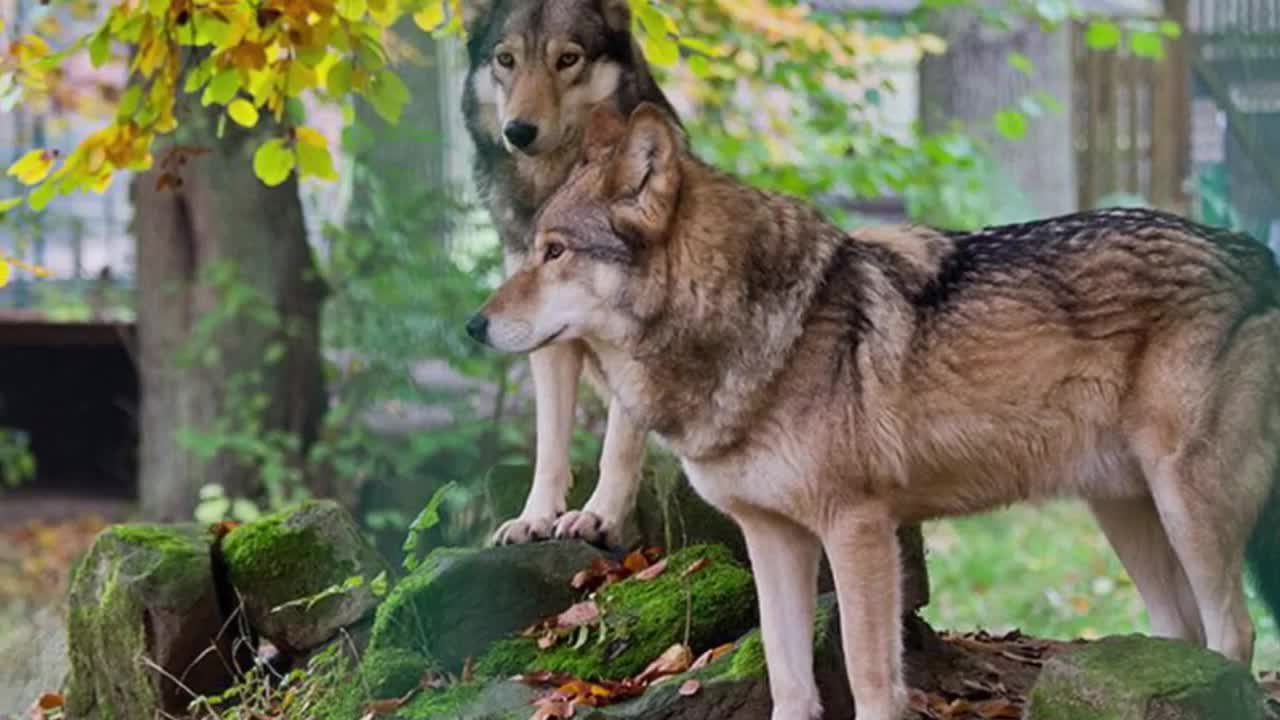}
    }
    \subfloat{
        \includegraphics[height=3.36cm]{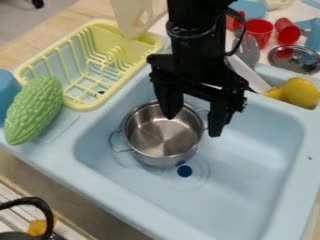}
    }
    \subfloat{
        \includegraphics[height=3.36cm]{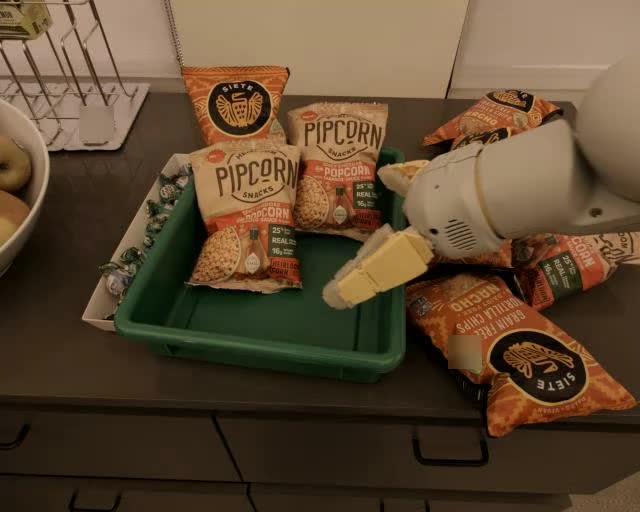}
    }
    \\
    \vspace{0.18cm}
    \subfloat{
        \includegraphics[height=3.0cm]{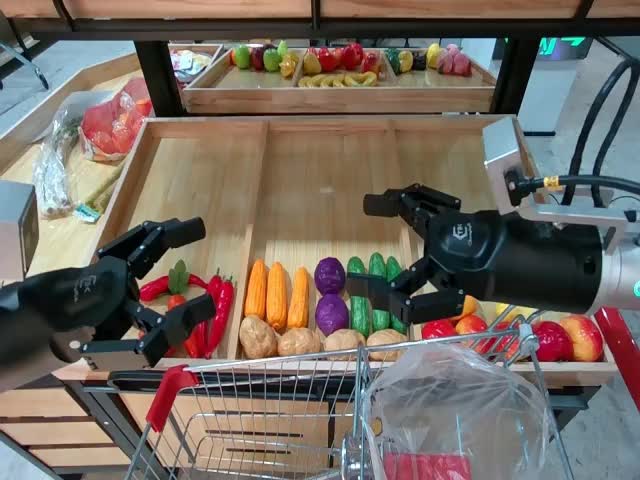}
    }
    \subfloat{
        \includegraphics[height=3.0cm]{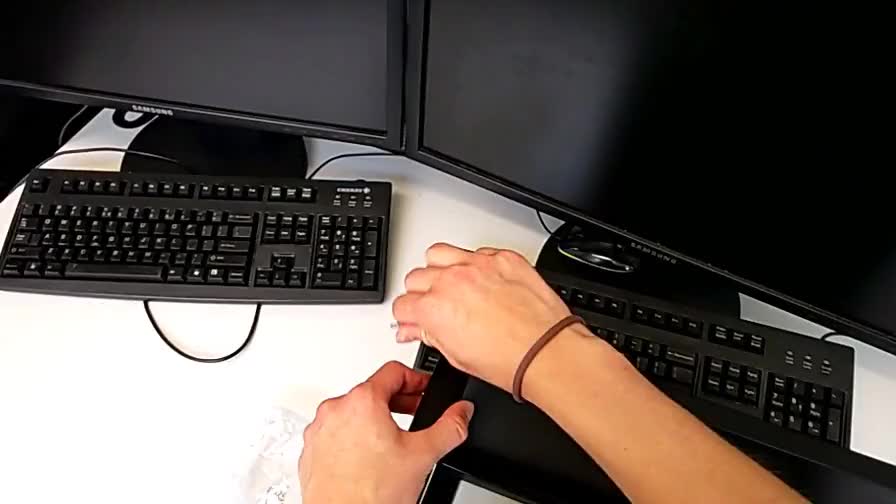}
    }
    \subfloat{
        \includegraphics[height=3cm]{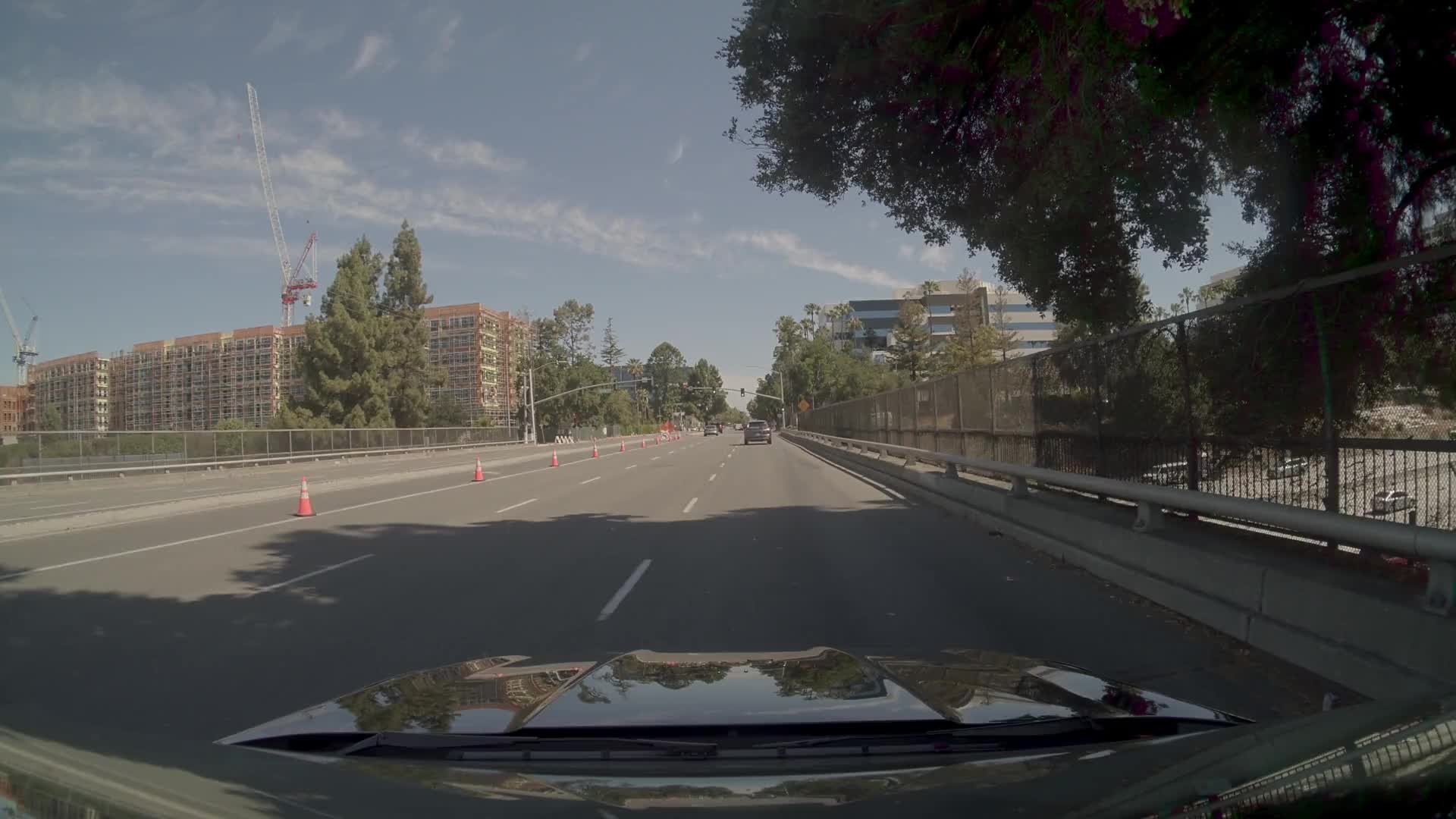}
    }
    \caption{Example of video frames from our Physical AI supervised fine-tuning datasets.}
    \label{fig:data_physical_ai_sft}
\end{figure}

For physical common sense, we build VQA datasets to answer free-form and multiple-choice questions from videos. For embodied reasoning, we subsample and convert existing datasets into our SFT datasets, where we cover a wide range of tasks across different embodiments, including humans, robot arms, humanoid robots, and autonomous vehicles. For each dataset, we collect two types of annotations: \textbf{understanding} and \textbf{reasoning}. An understanding annotation contains the video's question and answer for common sense and a detailed description of states and actions in the video (a structured video caption). A reasoning annotation contains a long chain-of-thought (CoT) thinking trace for the given text prompt and an input video.
Additionally, we further curate specific reasoning SFT datasets to enhance our models' ablitity to understand spatiotemporal visual stimuli (through puzzles and arrow of time in videos) as well as object permanence (through phyiscs-based simulation). These datasets are collectively referred to as intuitive physics. 
\cref{tab:data_physical_ai_sft} summarizes the datasets we used for Physical AI SFT and \cref{fig:data_physical_ai_sft} shows examples of video frames from the Physical AI SFT datasets.
Next, we describe the setting-specific curation pipelines in detail.

\begin{table}[b]
    \small
    \captionsetup{justification=centering}
    \caption{A summary of datasets used for physical AI supervised fine-tuning.}
    \centering
    \resizebox{\textwidth}{!}{
    \setlength{\tabcolsep}{6pt}
    \begin{tabular}{r|cc|ccccc|ccc|c}
    \toprule
         & \multicolumn{2}{c|}{Physical Common Sense VQA} & \multicolumn{5}{c|}{Embodied Reasoning} & \multicolumn{3}{c|}{Intuitive Physics}\\
         & Free-form & MCQ & BridgeData V2 & RoboVQA & Agibot & HoloAssist & AV & Puzzle & AoT & Object Permanence & Total \\
        \midrule
        Understanding & 99K & 1.2M & 129.2K & 218.5K & 19.4K & 136.3K & 12.4K & - & - & - & 1.81M \\
        Reasoning & 59.4K & 605.0K & 129.1K & 920.0K & 19.4K & 136.3K & 12.3K & 11.0K & 30.0K & 10.0k & 1.93M \\
    \bottomrule
    \end{tabular}}
    \label{tab:data_physical_ai_sft}
\end{table}

\subsubsection{Physical Common Sense SFT}

As stated earlier, for physical common sense, we collect VQA datasets that consist of both free-form and multiple-choice questions (MCQs). Our curation pipeline for physical common sense data consists of five stages:
\begin{enumerate}
    \item \textbf{Human-in-the-loop Video Curation.} We curate a set of high-quality videos based on human preferences. We extract short clips from these videos and use them as training samples.
    \item \textbf{Detailed Captioning.} We employ either reliable human annotators or pre-trained vision-language models (VLMs) to extract detailed descriptions of videos. These serve as ``captions'' which we use to construct understanding and reasoning annotations for video clips.
    \item \textbf{Curating QA Pairs.} We prompt an LLM to construct free-form or multiple-choice questions based on the detailed clip descriptions. We construct two kinds of questions: (1) ``understanding'' questions that cover the content in the video (as observed through the captions) and (2) hypothetical ``reasoning'' questions that require information from the caption for constructing the question but cannot be answered directly based on the detailed caption. The ``reasoning'' questions require more thought beyond just perceiving the events and objects in the clip. Our reasoning questions focus on common sense reasoning, spatial reasoning, and temporal reasoning from videos. Prompt~\ref{prompt:commonsense_reasoning_question_construction} shows a sample question-construction prompt template used to produce reasoning questions.
    \item \textbf{Extracting Reasoning Traces.} To obtain complete ``reasoning'' annotations, we prompt a DeepSeek-R1~\citep{deepseekai2025deepseekr1} to answer the reasoning subset of questions by using the detailed caption as context. Then, we parse DeepSeek-R1's response into a thinking trace and an answer. We find it is important to ask questions that cannot be directly answered from the caption. Otherwise, R1 can directly retrieve the answer from the provided caption, making the thinking trace invalid for model training. Our ``reasoning'' annotations consist of the reasoning questions, corresponding clips, thinking traces, and answers. Prompt~\ref{prompt:commonsense_reasoning_extraction} shows a sample prompt used to elicit reasoning from DeepSeek-R1.
    \item \textbf{Cleaning \& Rewriting.} Finally, we employ a rule-based cleaning and rewriting stage for the ``reasoning'' annotations to produce valid SFT samples. Since we compress the visual context of the clip into text, rewriting helps remove unwanted references such as ``description'' or ``caption'' in the SFT training samples.
\end{enumerate}

Using the aforementioned pipeline, we curated physical common sense VQA datasets consisting of both free-form and multiple-choice questions, with a few additional considerations as outlined below:

\noindent\textbf{Free-form Questions}: We used 9.9K videos from the curated set of ``high-quality'' clips and obtained corresponding detailed descriptions annotated by humans. The average length of human-annotated captions is $297.4 \pm 46.4$ words. For free-form questions, we obtain $\sim$99k understanding SFT samples and $\sim$59.4k reasoning SFT samples through the pipeline described above.

\noindent\textbf{Multiple Choice Questions (MCQs)}: To ensure our model is capable of answering multiple-choice questions (MCQs), we additionally collect a set of ``understanding'' and ``reasoning'' MCQs for the high-quality curated clips. Unlike the free-form questions, we first annotate the set of $\sim$1.2M high-quality clips with detailed captions from a VLM. Using these captions, we construct $\sim$2.4M ``understanding'' MCQs. Then, we take a subset of $\sim$356k clips and use the detailed captions to produce $\sim$600k ``reasoning MCQs''.

\subsubsection{Embodied Reasoning SFT}
\label{sub:embodied_reasoning_sft}
\begin{figure}[!htp]
\centering
\includegraphics[width=0.95\textwidth]{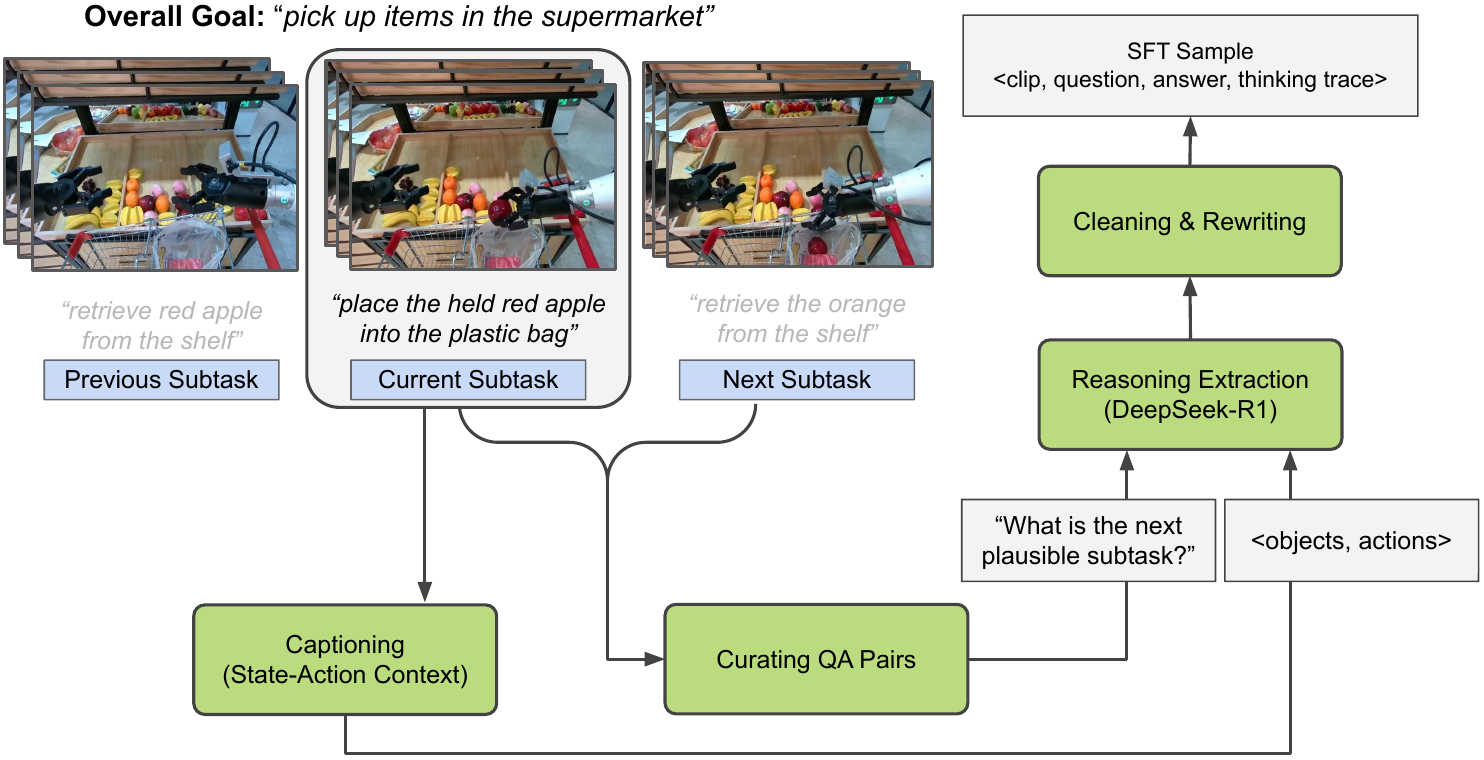}
\caption{Embodied reasoning SFT data curation pipeline. We demonstrate an illustrative example for AgiBot, where we (1) extract short horizon segments corresponding to the subtask, (2) caption the extracted clip to obtain state-action context, (3) curate QA pairs for ``next plausible subtask prediction'', (4) prompt R1 with the question and caption to elicit reasoning, (5) clean and rewrite the reasoning trace to obtain valid SFT samples.}
\label{fig:embodied_reasoning_data_pipeline}
\end{figure}

Our embodied reasoning SFT data-curation pipeline focuses on three key properties essential for decision-making in Physical AI agents: (1) ``task-completion verification'': the ability to determine whether a task or subtask has been successfully completed; (2) ``action affordance'': the ability to assess whether executing a specific action or making progress toward a goal is possible; and (3) ``next plausible action prediction'': the ability to identify the most plausible next action or subtask to advance toward a specified goal. These properties are fundamental for effective decision-making across various embodiments and task configurations. To develop these reasoning capabilities, we curate SFT samples from both public and proprietary datasets. The embodied reasoning SFT dataset we use contains structured entries with four components: visual captions, questions, corresponding answers, and detailed reasoning traces.

Videos demonstrating Physical AI agents pursuing goal-oriented tasks serve as our primary source of embodied reasoning data. We collect SFT data from sources featuring demonstrations of humans, robots, or vehicles executing specific tasks. For embodied reasoning, we focus specifically on short-horizon reasoning related to our key properties of interest --- determining whether an agent can reason about an immediate next subtask or action given a goal (affordance and next plausible action prediction), or evaluate the successful completion of short-horizon tasks (verifying task-completion). Since existing Physical AI demonstration datasets may lack the dense annotations needed to extract such localized action or subtask sequences, we use a series of specialized steps to extract such segments. We ensure our curated dataset is rich in terms of diversity, short-horizon granularity (immediate next action or immediate next subtask), embodiments and reasoning prompts. The curation pipeline we use has the following general steps (also illustrated in~\cref{fig:embodied_reasoning_data_pipeline}):

\begin{enumerate}
    \item \textbf{Extracting Short-Horizon Segments}. Since we are interested in short-horizon reasoning tasks, we break down long video demonstrations into concise clips focused on short-horizon reasoning tasks. These segments capture either individual actions (\eg, ``move left'') or distinct subtasks (\eg, ``open fridge door''). When existing datasets already provide appropriately segmented clips or timestamps, we utilize them directly. Otherwise, we leverage complementary annotations such as action primitives and plans to extract these short-horizon segments.
    \item \textbf{Annotating State-Action Context}. For every short-horizon clip, we use a VLM to produce structured captions that detail the present objects, their attributes and associated actions. When datasets provide supplementary annotations that can enhance the quality of such structured captions, we incorporate this information into the VLM prompt. These constitute our ``understanding'' annotations for embodied reasoning SFT. For AV data, we directly use human-annotated captions.
    \item \textbf{Curating Reasoning QA Pairs}. We develop reasoning question-answer pairs focusing on our key properties of interest, based on available subtask and action annotations. For datasets already containing suitable QA pairs addressing our target properties, we apply minimal rule-based preprocessing before adding them to our pool of embodied reasoning SFT data.
    \item \textbf{Extracting Reasoning Traces}. We utilize DeepSeek-R1~\citep{deepseekai2025deepseekr1} to generate reasoning traces for our curated QA pairs. Since R1 lacks visual processing capabilities, we construct prompts containing the state-action context, question, and additional information (such as subtask instructions or overall goals) to elicit appropriate reasoning traces.~\cref{fig:embodied_reasoning_data_pipeline} demonstrates this process, while Prompt~\ref{prompt:agibot_reasoning} shows an example user prompt that transforms visual information into textual context for a short-horizon question about ``next plausible action''.
    \item \textbf{Cleaning \& Rewriting}. Finally, we use rule-based cleaning and rewriting to retain only valid and useful reasoning traces. Since we compress the visual context of the clip into text, rewriting helps remove unwanted references to the ``description'' or ``caption''.
\end{enumerate}

The exact specifics of each step in the curation pipeline vary slightly across datasets, but the overall pipeline remains the same. We now describe how this pipeline is applied across the individual data sources.

\noindent \textbf{BridgeData V2}: 
BridgeData V2~\citep{walke2023bridgedata} is designed to advance scalable robot learning by providing a wide array of robotic manipulation behaviors. The dataset emphasizes foundational object manipulation tasks such as pick-and-place, pushing, and sweeping, alongside more complex activities like stacking blocks and folding cloths. It comprises 60,096 trajectories, including 50,365 teleoperated demonstrations and 9,731 scripted pick-and-place rollouts, spanning 13 distinct skills across 24 diverse environments. 
Each trajectory is annotated with natural language instructions corresponding to the task performed by the robot. The environments are categorized into four groups (toy kitchens, tabletops, toy sinks, and other), with a significant portion of the data collected from seven unique toy kitchens featuring combinations of sinks, stoves, and microwaves. 
We first split the videos from the dataset ``train'' split and obtain 129.5K video clips. Then we use a VLM to caption the video clips as the understanding annotations. In the captioning prompt, we also provide additional information such as detected objects and action sequences from ECoT~\citep{zawalski2024ecot}. We produce only ``next plausible action prediction'' reasoning QA pairs for BridgeData V2, where the answers correspond to action primitives such as \textit{move left}. The reasoning annotations are generated by feeding the captions and questions to DeepSeek-R1.

\noindent \textbf{RoboVQA}: 
RoboVQA~\citep{sermanet2024robovqa} is a large-scale robotics-focused visual question answering dataset. It consists of videos, instructions, and question-answer pairs of agents (robots, humans, humans-with-grasping-tools) executing a task. RoboVQA has 6 different question-types that cover aspects related to planning, verifying task-completion, discriminative affordance, generative affordance, past description and future prediction (these correspond to properties outlined before). We directly use the clips in RoboVQA without any clipping to obtain a dataset of $\sim$220k clips. We caption these clips using the VLM and extract reasoning traces from DeepSeek-R1 by combining the task-context, caption and question into a suitable user prompt. This leads to $\sim$930k QA pairs with reasoning traces. We filter a suitable subset post-cleaning and use the clips \& QA pairs in the ``train'' split of the dataset for SFT. SFT samples from RoboVQA encompass all the 3 desired properties in our embodied reasoning curation pipeline.

\noindent \textbf{AgiBot}: 
AgiBot World \citep{contributors2024agibotworldrepo} is a high-fidelity robot manipulation dataset. The data is collected using the AgiBot G1 hardware platform, covering a wide range of real-life tasks. It consists of 36 tasks. Each task contains multiple episodes that vary in terms of environment and objects. We subsample a portion of episodes for each task, resulting in a total of 3,300 videos. Each video is annotated with overall task information and multiple subtask annotations, including start and end frames. We utilize these action annotations to split the videos into clips, resulting in a final dataset of 19.8K clips. These clips are captioned by the VLM to convert the visual information to the scene/object descriptions and their movements. 
We produce only ``next plausible subtask prediction'' questions for AgiBot, where the answer corresponds to a subtask (``\textit{place cucumber in the bag}'').
Then we use DeepSeek-R1 to reason about the next possible subtask required to complete the task, based on the generated captions.

\noindent \textbf{HoloAssist}: 
Egocentric datasets capture crucial first-person perspectives that provide natural and immersive understanding of human actions, but present unique challenges including camera motion, subtle movements, occlusions, out-of-view objects, spatial perspective issues, and the need for global scene comprehension. Despite these challenges, they remain valuable for developing embodied decision-making capabilities in Physical AI systems, potentially enabling human-like interpretation and response to real-world environments. We choose to build upon HoloAssist \citep{HoloAssist2023}, which contains 166 hours of egocentric video focused on object-centric manipulation tasks. Notably, HoloAssist uniquely includes human mistakes and the corrective steps taken to fix them. These insights can help Physical AI learn in a way that mirrors how humans learn and refine their understanding with objects in the real world. Using timestamped coarse- and fine-grained action annotations from HoloAssist, we split 1,758 videos into a final dataset of 139,653 clips. We employ a VLM to generate caption annotations. We produce only ``next plausible subtask prediction'' questions for HoloAssist, where the answer corresponds to a subtask. We use DeepSeek-R1 to produce reasoning traces for predicting the next possible subtasks needed to complete a task based on the generated captions. In each pipeline, we provide the task annotation as the overall goal and the fine-grained annotation as the current subtask to supplement captioning.

\noindent \textbf{Autonomous Vehicles (AV)}: 
As a key domain in Physical AI, autonomous vehicles (AV) rely on large-scale and high-quality data to enable safe and reliable self-driving experiences, particularly in the era of rapidly scaling end-to-end systems. In this work, to avoid captioning hallucinations --- especially in subtle behaviors and complex interactions --- we utilize proprietary datasets with high-quality captions annotated by humans. Our dataset consists of $\sim$12.4K videos, each 20 seconds long, totaling around 70 hours.
Each caption includes three categories: (1) \textit{general description}, which details ego behaviors, environmental conditions (e.g., scene type, time of day, weather, road conditions), and critical objects (e.g., vehicles, pedestrians, cyclists, traffic lights, traffic signs); (2) \textit{driving difficulty}, which provides a concise assessment of driving complexity based on the level of driver attention required and the scenario's uniqueness or risk; and (3) \textit{notice}, which highlights notable events, such as signs and signals, road user interactions, and abnormal behaviors. By leveraging these captions, we transform driving videos into structured descriptions. DeepSeek-R1 is then used to generate reasoning traces to predict the next most likely immediate action the ego vehicle will be taking based on these structured descriptions.

\subsubsection{Intuitive Physics SFT: Spatial Puzzles, Arrow-of-Time and Object Permanence}
\label{sub:physical_ai_sft_intuitive_physics}
While the prior SFT stages enable domain-specific reasoning for Physical AI applications, we incorporate additional SFT stages to develop fundamental reasoning abilities centered around intuitive physics. Although intuitive physics reasoning capabilities encompass a broad taxonomy 
(see~\cref{tab:common_sense_ontology}), we focus specifically on three key aspects: reasoning about spatial continuity (through spatial puzzles), reasoning about the arrow of time (through the temporal order of events in videos), and reasoning about objectness (through simulation-based settings that assess object permanence). These tasks are inherently self-supervised by construction, which simplifies the data curation process. Despite significant progress in more sophisticated tasks, current state-of-the-art VLMs still underperform substantially on these simpler fundamental reasoning objectives. To address these limitations, we curate specialized SFT datasets targeting spatial continuity, arrow of time, and object permanence.

\noindent \textbf{Reasoning about Spatial Continuity: Spatial Puzzles.}
In addition to spatial relationships, understanding spatial continuity is crucial for Physical AI tasks. To imbue our models with a fundamental understanding of spatial continuity, we further finetune them on a task of solving spatial puzzles. Specifically, we curate 3000 videos featuring diverse backgrounds, motions, and camera poses. For each video, we extract the first frame and divide it into 2$\times$2 patches. These patches are then shuffled to create a new clip, where one frame is one patch. We prompt the model to identify the left, top, bottom, and right positions relative to the original frame. To further increase the complexity of the task, we introduce seven additional distractor images, each also divided into 2$\times$2 patches. This results in a total of 32 shuffled frames for a single sample, which are provided to the model to reason about the correct positions. Additionally, we design a few ``identity''-driven supplementary tasks --- determining which two or three frames originate from the same image as the first frame. Akin to contrastive learning, this task requires models to develop strong spatial reasoning capabilities while distinguishing between relevant and irrelevant samples, except now with reasoning.

To generate high-quality reasoning data for spatial continuity, we first caption each of the 32 patches using a VLM and input these descriptions into DeepSeek-R1 to solve one of the three tasks. 
We only retain samples where R1 makes the correct prediction. Each of the 3,000 images is processed multiple times with varying distractors and shuffle orders. After filtering, our final dataset consists of 11k videos. Prompt~\ref{prompt:puzzle_reasoning_prompt} shows a sample prompt to elicit reasoning traces.

\noindent \textbf{Reasoning about Time: Arrow-of-Time (AoT).} Similar to space, we also imbue our models with the ability to reason about time, specifically the temporal order of events in a macroscopic scale. We want our model to understand that time is inherently irreversible at a macroscopic scale, and the same can be perceived through motion and activity patterns in videos. Being able to reason about the one-way arrow of time is crucial for Physical AI, as it is closely linked to fundamental physical phenomena such as entropy, gravity, and causality. Additionally, it can serve as a proxy supervisory signal for learning intuitive physics. In particular, temporally altered or reversed videos contain physics-defying artifacts that a Physical AI capable reasoning model should be able to evaluate.

We construct an SFT dataset containing 30,000 short video clips and their reversed versions using a subset of videos from the training dataset of Cosmos-Predict1~\citep{agarwal2025cosmos}. This subset features videos containing diverse and complex tasks that frequently involve significant motion. We prioritize videos with large motion, as they serve as the most representative examples for distinguishing the arrow of time. Unlike the prior stages, we use a VLM to directly extract a reasoning trace for both forward and reverse playback clips. We found that applying the same procedure as \Cref{sub:embodied_reasoning_sft} led to suboptimal results for thinking trace extraction using R1. To improve curation, we explicitly indicate whether the video is played forward or in reverse in the user prompt, helping the VLM generate more reasonable justifications. Additionally, we carefully design the prompts to ensure that reasoning traces maintain a consistent style and length for both playback directions. Prompt.~\ref{prompt:aot_reasoning_prompt} shows an example prompt template used for this purpose. To encourage diversity in the reasoning, we curate two distinct reasoning traces for each forward and backward video.

\noindent \textbf{Reasoning about Objectness: Object Permanence.} While spatial continuity and temporal order provide foundations for understanding physical relationships and sequences, object permanence --- the understanding that objects continue to exist even when they cannot be directly observed --- represents a critical reasoning capability that is fundamental for Physical AI agents. Without strong object permanence reasoning, VLMs would struggle with even basic real-world scenarios where objects frequently move in and out of view or become occluded, severely limiting their utility in Physical AI applications that require consistent object tracking and prediction capabilities.

For object permanence, we construct an SFT dataset containing 10K clips synthesized by a robot simulation platform, Libero \citep{liu2023libero}. Libero offers 130 robot arm object manipulation tasks across diverse environments, desktop objects and pre-recorded arm action sequences. To enhance scene diversity, we randomly sample setups from these tasks and apply object permutation and perturbation. The camera is positioned to face the table center and orbits around the scene by selecting a random start and end point on a sphere. During playback of the pre-recorded arm actions, the camera smoothly interpolates from the start to the end point before returning near its starting position. Throughout this transition, some objects may be temporarily occluded, and once fully occluded, certain objects may be randomly removed from the scene. We prompt the model with appropriate context and ask it to analyze each clip and determine whether any objects disappear unexpectedly, violating object permanence. To ensure the model generates reasoning traces that consistently lead to the correct answer, we include hints in the prompt to indicate which objects disappear and do not reappear. However, in the final SFT dataset, these hints are removed from the prompts. For object permanence, we find that the standard pipeline of compressing visual context into captions is suboptimal for extracting useful reasoning traces for SFT. To address this, we extract thinking traces from an intermediate version of Cosmos-Reason1-7B. Prompt~\ref{prompt:permanence_reasoning_prompt} shows a sample prompt used to elicit a reasoning trace from Cosmos-Reason1-7B.

\subsection{Physical AI Reinforcement Learning}
\label{sec::physicalai_rl}

\begin{table}[htbp]
    \small
    \captionsetup{justification=centering}
    \caption{
        Datasets for Physical AI reinforcement learning post-training.
    }
    \centering
    \resizebox{\textwidth}{!}{
    \setlength{\tabcolsep}{6pt}
    \begin{tabular}{r|c|ccccc|ccc|c}
    \toprule
         & \multicolumn{1}{c|}{Common Sense} & \multicolumn{5}{c|}{Embodied Reasoning} & \multicolumn{3}{c|}{Intuitive Physics} \\
         & MCQ & BridgeData V2 & RoboVQA & Agibot & HoloAssist & AV & Puzzle & AoT & Object Permanence & Total\\
        \midrule
        Reasoning & 5{,}133 & 240 & 252 & 200 & 200 & 200 & 3{,}998 & 9{,}994 & 10{,}087 & 30{,}304\\
    \bottomrule
    \end{tabular}}
    \label{tab:data_physical_ai_rl}
\end{table}

While fine-tuning establishes foundational physical common-sense and embodied reasoning capabilities, we further enhance these abilities through reinforcement learning post-training. This approach requires effective reward mechanisms, which we implement using tried-and-tested rule-based and verifiable rewards following \citet{deepseekai2025deepseekr1}. Unlike LLM domains such as mathematics and coding --- where correct answers and formats are precisely defined --- physical common sense and embodied reasoning typically involve free-form, open-ended responses that complicate reward assignment.

To address this challenge, we convert samples from our reasoning SFT data sources (excluding thinking traces) into multiple-choice questions with single correct answers (see Table.~\ref{tab:data_physical_ai_rl} for the number of samples). This transformation inherently enables simple, rule-based verification of responses. Our RL post-training dataset incorporates samples from all Physical AI SFT data sources, with certain subsets --- specifically Spatial Puzzles, AoT, and Object Permanence data --- already existing in binary question format, making them directly suitable as MCQs without modification. We manually verify the quality of samples used for RL post-training. We detail the specific characteristic of individual data sources below:

\noindent \textbf{Physical Common Sense RL Data.}
We collect 5133 human annotated binary and multiple-choice questions from 1989 videos. To help control the difficulty of the questions, we use the annotated questions to evaluate four models including GPT-4o, Gemini Flason 2.0, Qwen2.5-VL-7B, and our \cosmosreason-7B model. Based on the evaluation results, we further divided the collected data into two subsets: (1) the easy subset with questions that all models got them correct; (2) the hard subset with questions that at least one model got them wrong.

\noindent \textbf{Embodied Reasoning RL Data.} We select 200-250 SFT samples from each embodied reasoning data source and convert them to multiple-choice questions (MCQs). To ensure high-quality RL post-training, we carefully verify that these samples are free from answer and instruction ambiguity while maintaining balanced distribution across MCQ options to prevent potential reward hacking. This process requires some manual intervention, particularly for non-binary questions where we must select appropriate distractor options that are plausible yet clearly incorrect. The human-in-the-loop is required to ensure question quality, making it difficult to generate large-scale MCQ data for training.

\noindent \textbf{Intuitive Physics RL Data.} As previously described, our self-supervised intuitive physics SFT data naturally exists in MCQ format by design, making it scalable to generate diverse questions. For these tasks, we implement additional quality assurance measures to ensure balanced option distributions across all samples. We carefully avoid any overlap with clips used during SFT to prevent early saturation during RL post-training. For the RL post-training phase, we curate 24079 high-quality samples across spatial continuity, arrow of time, and object permanence tasks.

\section{Benchmark}
\label{sec::benchmark}

We compare our trained models with other counterparts on a benchmark geared specifically towards measuring reasoning capabilities about physical common sense, embodied decision-making.
In this section, we discuss the procedure of building our common sense and embodied reasoning benchmarks (\Cref{tab:benchmark_stat}).
We evaluate models by asking either binary yes / no questions or multiple-choice questions (MCQs) based on video context. We note that our benchmarks necessitate reasoning by construction (to arrive at the correct answer) and we only measure the accuracy of the final answer. We leave quantitatively assessing the quality of the thinking trace for future work.

\begin{table}[h]
    \small
    \captionsetup{justification=centering}
    \caption{
        Statistics of our curated benchmarks.
    }
    \centering
    \resizebox{0.8\textwidth}{!}{
    \setlength{\tabcolsep}{6pt}
    \begin{tabular}{c|cccccc|c}
    \toprule
         \multicolumn{1}{c|}{Common Sense} & \multicolumn{6}{c|}{Embodied Reasoning}  \\
         MCQ & BridgeData V2 & RoboVQA & RoboFail & Agibot & HoloAssist & AV & Total\\
        \midrule
        604 & 100 & 110 & 100 & 100 & 100 & 100 & 1214 \\

    \bottomrule
    \end{tabular}}
    \label{tab:benchmark_stat}
\end{table}

\subsection{Physical Common Sense Reasoning}
\label{sec::benchmark_physical_reasoning}

\begin{figure}[!htp]
    \centering
    \includegraphics[width=0.95\textwidth]{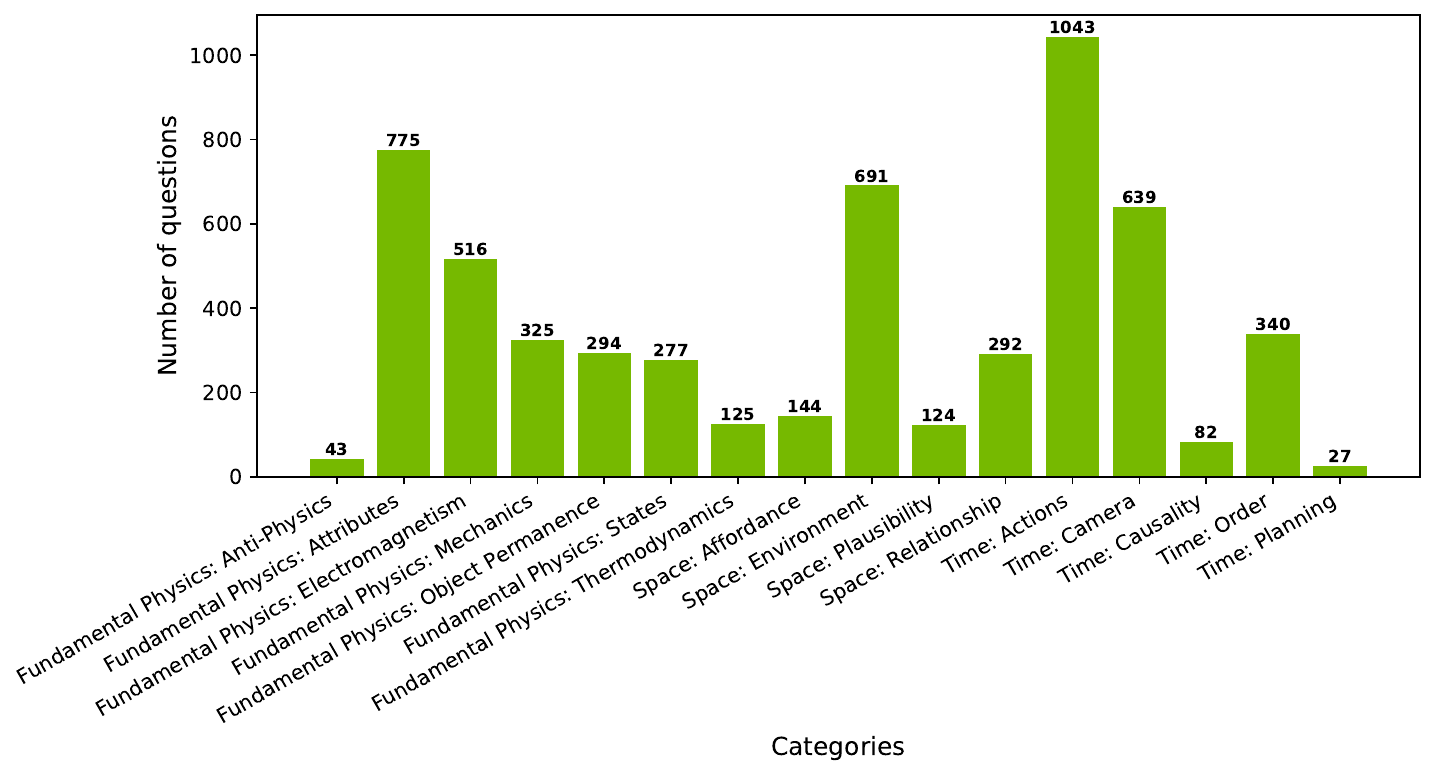}
    \caption{An illustration showing the categorical distribution of initial questions collected for physical common sense benchmark according to the ontology defined in~\cref{tab:common_sense_ontology}. We select a subset of 604 questions as our evaluation benchmark.}
    \label{fig:benchmark_common_sense}
\end{figure}

We construct a physical common sense reasoning benchmark by manually curating questions about internet video clips according to the ontology defined in~\cref{sec::definition_common_sense}.
We initially collected a pool of 5737 questions, including 2828 binary questions and 2909 multiple-choice questions.~\cref{fig:benchmark_common_sense} shows the distribution of question categories according to our ontology.
After that, we went through a manual process to carefully select a subset of 604 questions from 426 video clips used as our physical common sense benchmark, from which 336 are binary questions and 268 are multiple choice questions.
Among the 604 questions, 80 ($13.25\%$) are about Space, 298 ($49.33\%$) are about Time, and 226 ($37.4\%$) are about Fundamental Physics.

\subsection{Embodied Reasoning}
\label{sec::benchmark_embodied_reasoning}

Similar to our SFT data-curation pipeline for embodied reasoning, we constrain our embodied reasoning benchmark to focus on the previously outlined properties before ---  (1) ``task-completion verification”: the ability to determine whether a task or subtask has been successfully completed; (2) ``action affordance”: the ability to assess whether executing a specific action or making progress toward a goal is possible; and (3) ``next plausible action prediction”: the ability to identify the most plausible next action or subtask to advance toward a specified goal. We present our embodied reasoning benchmark samples as multiple-choice questions (MCQs) to enable automatic evaluation across models. We adopt a few key steps to ensure our benchmark is useful for measuring embodied reasoning abilities.

\begin{enumerate}
    \item \textbf{Unified Question Templates.} We adopt a unified format for question formulation to ensure that reasoning is conditioned on visual input rather than textual cues. This approach also helps align action granularity and reduce ambiguity across different datasets.
    \item \textbf{Unified Action Granularity.} We pay special attention to action granularity. In predicting the next immediate action, multiple choices could be potentially correct. For example, the action ``water the plant'' may involve steps such as ``grab the watering can'', ``move the watering can'', and ``pour the watering can''. However, these steps can also be broken down into finer sub-actions like ``move left'' or ``tilt down''. To address this complexity, we use a hierarchy of actions~\citep{belkhale2024rt}: we define atomic-level actions as ``actions'', more coarse-grained actions as ``subtasks'', and dataset-specific tasks as ``goals''.
    \item \textbf{Manual Refinement.} Apart from these systemic approaches to addressing ambiguity, we also manually refine the MCQ choices. These modifications help resolve overly similar options, enforce visual reasoning by preventing answers from being inferred solely from text, and emphasize the full context of the entire clip rather than just the earlier frames.
\end{enumerate}

\noindent\textbf{RoboVQA}: 
We sample 101 clips from the ``val'' split (excluded from SFT training) of the processed RoboVQA samples for our benchmark. For benchmarking purposes, we only consider clips and question-answer pairs that correspond to ether verifying task-completion (whether a provided instruction was followed successfully) or affordance (whether it is possible to follow an instruction / complete a task). These are included as multiple-choice yes / no questions.

\noindent\textbf{RoboFail}: 
In addition to RoboVQA, we also manually curate and annotate 100 examples from the RoboFail~\citep{liu2023reflect} dataset to create a harder ``action affordance'' and ``task completion verification'' evaluation split, where the hardness of samples is dictated by (1) the necessity of highly observant perception or comprehensive temporal context processing, (2) identifying physical constraints blocking follow-through for an action (unlike RoboVQA where it's impossible to complete an action due to perception mismatches or irrelevant instructions) and (3) being able to reason about nuanced questions.

\noindent\textbf{BridgeData V2}:
We split the videos from ``val'' split of BridgeData V2 into clips following the same way as the training set. Then we sample 100 clips to create 100 multiple-choice QA pairs as the benchmark. In each question, we provide the task that the robot gripper is instructed to perform in the clip, and ask what the most plausible next immediate action is given what the robot has done in the video.

\noindent\textbf{AgiBot}:
We sample 100 clips from the processed AgiBot SFT data to generate 100 multiple-choice QA pairs. For each clip, we additionally provide the task information and ask which of the given subtasks is the most likely next subtask the robot should work towards. We randomly sample the choices from the subtask sequence of the clip's entire trajectory. Notably, these trajectories are excluded from the training set.

\noindent\textbf{HoloAssist}:
We sample 100 clips from the processed HoloAssist SFT data to generate 100 multiple-choice QA pairs. For each clip, we additionally provide the coarse-grained action annotation as the overall goal and ask which of the given subtasks is the most likely next subtask. We randomly sample the choices from other fine-grained action annotations under that coarse-grained action. All 34 videos (out of a total of 1758) that contain these clips are excluded from the training set to prevent episode leakage. 

\noindent\textbf{AV}:
We curate 100 videos from the proprietary data to construct 100 multiple-choice QA pairs. These videos exhibit diverse lateral and longitudinal behaviors, along with rich interactions. The questions are designed to (1) predict the next immediate action the ego vehicle will most likely be taking, (2) verify the completion of a previously executed action, and (3) assess the possibility of a specific action in a given scenario. Note that these samples in the benchmark are excluded from the training set.

\section{Experiments}
\label{sec::experiment}

In this section, we illustrate the experiment setup for physical AI supervised fine-tuning and physical AI reinforcement learning of \cosmosreason and discuss the evaluation results on our benchmarks.
\subsection{Physical AI Supervised Fine-Tuning}
\label{sec::physical_ai_sft}

We train \cosmosreason-7B for $12.5K$ iterations with a cosine annealing learning rate decaying from $1\times 10^{-5}$ to $1\times 10^{-6}$. We train \cosmosreason-56B for $30K$ iterations with a learning rate of $1\times 10^{-5}$, followed by $20K$ iterations with a decayed learning rate of $1\times 10^{-6}$.
We use a global batch size of 256 for \cosmosreason-7B and 32 for \cosmosreason-56B. The fused Adam optimizer with $\beta_1, \beta_2 = (0.9, 0.95)$ and weight decay of $0.1$. We follow a balanced data sampling strategy during training so that no specific domain of interest is overrepresented during SFT.
Unless stated otherwise, for our models, we report the average accuracy of 5 inferences (temperature 0.6 and top-p 0.95) with different random seeds.
For evaluating other models, we employ a zero-shot chain-of-thought prompting~\citep{kojima2022large} by either calling their APIs (GPT-4o, OpenAI o1, Gemini 2.0 Flash) or using their open-sourced model checkpoints (Qwen2.5-VL).

\subsubsection{Physical Common Sense Results}
\label{sec::sft_common_sense}

\Cref{tab:exp_common_sense} shows the evaluation results on physical common sense benchmark. \cosmosreason-7B and \cosmosreason-56B demonstrate significantly improved capabilities on the physical common sense benchmark compared to their respective backbones, with the 56B variant achieving the best accuracy, slightly outperforming OpenAI o1.
These results highlight the effectiveness of our curated common sense dataset, laying a strong foundation for further RL improvements.

\begin{table}[h]
    \setlength{\tabcolsep}{4.7pt}
    \small
    \captionsetup{justification=centering}
    \caption{
        Evaluation on physical common sense benchmark.
    }
    \centering
    \setlength{\tabcolsep}{6pt}
    \begin{tabular}{rcccc}
    \toprule
         Methods &  Space & Time & Other Physics & Avg. \\
        \midrule
        Gemini 2.0 Flash & 53.8 & 50.0 & 46.9 & 50.2 \\
        GPT-4o & 61.3 & 54.7 & 50.9 & 55.6 \\
        OpenAI o1 & 63.8 & 58.1 & 58.0 & 59.9 \\
        \midrule
        Qwen2.5-VL-7B & 48.8 & 56.4 & 37.2 & 47.4 \\
        Nemotron-H-56B & 61.3 & 68.1 & 45.1 & 58.2 \\
        \midrule
        \cosmosreason-7B & 54.2& 58.7 & 50.0 & 54.3~\plus{6.9}\\
        \cosmosreason-56B & 61.3 & 65.5 & 53.9 & 60.2~\plus{2.0} \\
    \bottomrule
    \end{tabular}
    \label{tab:exp_common_sense}
\end{table}

\subsubsection{Embodied Reasoning Results}
\label{sec::sft_embodied_reasoning}

\Cref{tab:exp_embodied_reasoning} shows the evaluation results on the embodied reasoning benchmark. \cosmosreason models achieve significantly stronger results than the baseline models on this benchmark, with both the 7B and 56B variants demonstrating over a 10\% improvement compared to their respective backbone VLMs. The results demonstrate that our physical AI SFT is highly effective in boosting models' Physical AI embodied reasoning capabilities.

\begin{table}[h]
    \setlength{\tabcolsep}{4.7pt}
    \small
    \captionsetup{justification=centering}
    \caption{
        Evaluation on embodied reasoning benchmark.
    }
    \centering
    \setlength{\tabcolsep}{6pt}
    \begin{tabularx}{\textwidth}{>{\raggedright\arraybackslash}X*{7}{c}}
    \toprule
        Models & BridgeData V2 & RoboVQA & Agibot & HoloAssist & AV & RoboFail&Avg. \\
        \midrule
        Gemini 2.0 Flash & 25.0 & 78.2 & 29.0 & 44.0 & 37.0 & 67.0 & 46.7 \\
        GPT-4o & 42.0 & 71.8 & 32.0 & 65.0 & 46.0 & 63.0 & 53.3 \\
        OpenAI o1 & 42.0 & 80.0 & 44.0 & 63.0 & 37.0 & 61.0 & 54.5 \\
        \midrule
        Qwen2.5-VL-7B & 38.0	& 	82.5 & 40.4  & 	50.0 &	36.0 & 57.6  & 	50.8 \\
        Nemotron-H-56B & 37.0 & 77.2 & 37.0 & 65.0 & 41.0 & 64.0 & 53.5 \\
        \midrule
        \cosmosreason-7B & 58.8 & 83.8 & 49.4	& 63.0	& 55.6 &		 60.0 &		61.8~\plus{11.0} \\
        \cosmosreason-56B & 65.0 & 80.0 & 47.6 & 57.8 & 65.8 & 66.2 & 63.7~\plus{10.2} \\
    \bottomrule
    \end{tabularx}
    \label{tab:exp_embodied_reasoning}
\end{table}

\subsubsection{Intuitive Physics Results}
\label{sec::sft_intuitive_physics}

Although VLMs are often perceived as specialists achieving superhuman performance, our study reveals that many struggle with basic physics reasoning.
To test the model's capacity to understand intuitive physics, we curate 100 videos for each of the three tasks: the arrow of time, spatial puzzle, and object permanence, and generate 100 questions following pipeline in \Cref{sub:physical_ai_sft_intuitive_physics}. We conduct data decontamination to ensure no overlap with the training data. We evaluate model performance on the curated test set.

\Cref{tab:exp_rl_ssl} shows that the existing models struggle to perform above chance level on arrow of time and object permanence tasks. Notably, GPT-4o and OpenAI o1 handle spatial puzzles much better than random guessing. This observation suggests that current multimodal models are more proficient at reasoning about spatial relationships than temporal dynamics. Given that these models generally perform well on standard benchmarks like MMMU, this suggests that existing evaluations fail to capture their understanding of the physical world.
However, our curated intuitive physics dataset enables the 7B model to improve significantly across all three tasks, demonstrating \cosmosreason's fundamental capability to reason in intuitive physics.
\subsection{Physical AI Reinforcement Learning}
\label{sec::exp_rl}

 We post-train \cosmosreason with simple, rule-based verifiable rewards to further enhance their Physical AI reasoning abilities. We summarize our experimental findings on physical common sense, embodied and intuitive physics reasoning tasks.

\subsubsection{Experiment Setup}

We adopt two types of rule-based rewards to optimize our model towards accurate reasoning of Physical AI:
\begin{enumerate}
    \item An \textbf{accuracy reward} evaluates whether the model's response, enclosed within the <answer></answer> tags, matches the ground truth. Since we exclusively use MCQs for RL, this verification can be performed simply through string matching.
    \item A \textbf{format reward} encourages the model to encapsulate the thinking process in the <think></think> tag, and the answer in the <answer></answer> tag. This is implemented as a regular expression matching.
\end{enumerate}

During training, we sample from each RL dataset with equal probability, ensuring balanced representation across different domains. We also dynamically shuffle the MCQ choices on-the-fly to encourage generalization. We use a global batch size of 128 questions, for each question we sample 9 outputs, each with a maximum length cutoff of 6144 tokens. We set the learning rate to $4\times 10^{-6}$, the coefficient of the KL penalty term to $0.005$, and train the model for 500 iterations.

\begin{table}[h]
    \small
    \captionsetup{justification=centering}
    \caption{
        Evaluation on physical common sense and embodied reasoning benchmark.
    }
    \centering
    \setlength{\tabcolsep}{3pt} 
    \begin{tabular}{l*{8}{c}} 
    \toprule
        Models & Common Sense & BridgeData V2 & RoboVQA & Agibot & HoloAssist & AV & RoboFail & Avg. \\
    \midrule
        Cosmos-Reason1-7B & 54.3 & 58.8 & 83.8 & 49.4	& 63.0	& 55.6 &		 60.0 &	60.7 \\
        \textbf{+ Physical AI RL} & 56.2 & 73.5 & 86.8 & 54.2 & 60.0 & 67.0 & 62.0 & 65.7~\plus{5.0} \\
    \bottomrule
    \end{tabular}
    \label{tab:exp_rl}
\end{table}
\begin{table}[h]
    \small
    \captionsetup{justification=centering}
    \caption{
        Evaluation on intuitive physics benchmark.
    }
    \centering
    \setlength{\tabcolsep}{3pt} 
    \begin{tabular}{l*{4}{c}} 
    \toprule
        Models & Arrow of Time & Spatial Puzzle & Object Permanence & Avg. \\
    \midrule
        Random Guess & 50.0 & 25.0 & 50.0 & 41.7 \\
        \midrule
        Gemini 2.0 Flash& 50.0 & 31.0 & 48.0 & 43.0\\
        GPT-4o& 50.0 & 77.0 & 48.0 & 58.3 \\
        OpenAI o1& 51.0 & 64.0 & 49.0 &54.7\\
        \midrule
        Qwen2.5-VL-7B & 50.2	&  27.2	& 48.8	& 42.1\\
        Cosmos-Reason1-7B  & 56.0 & 85.4 & 82.0 & 74.5~\plus{32.4} \\
        \textbf{+ Physical AI RL} & 64.5 & 94.0 & 86.0 & 81.5~\plus{7.0} \\
    \bottomrule
    \end{tabular}
    \label{tab:exp_rl_ssl}
\end{table}

\subsubsection{Physical Common Sense and Embodied Reasoning Results}
\label{sec::rl_embodied_reasoning}

For physical common sense and embodied reasoning, we find that Physical AI RL post-training improves performance across most benchmark components, with the notable exception of RoboFail. 
The results are summarized in \Cref{tab:exp_rl}.
Performance on RoboFail remains consistently challenging through both SFT and RL stages. This is unsurprising given RoboFail's deliberate design as a hand-curated benchmark featuring difficult real-world scenarios testing ``action affordance'' and ``task-completion verification''. The benchmark's difficulty stems from several factors: (1) samples requiring highly observant perception or comprehensive temporal context processing, and (2) affordance questions involving complex physical constraints in action execution, unlike those in RoboVQA.

We attribute the stagnant performance on RoboFail primarily to insufficient representative training data. This hypothesis is supported by examining the specific error patterns in both finetuned and post-trained models, which include: inadequate reasoning about complex affordance scenarios, failure to be sufficiently observant of crucial visual details, and overthinking when encountering nuanced out-of-distribution questions. We believe targeted training on similar representative samples would significantly improve performance on this challenging benchmark. We deliberately maintain RoboFail in our evaluation suite as a valuable indicator of areas requiring further development in embodied reasoning capabilities.

\begin{figure}[p]
\centering
    
\begin{tcolorbox}[colback=nvidiagreen!3,colframe=nvidiagreen!75!white,title=\textsc{Embodied Reasoning Benchmark},left=0.5ex,right=0.5ex,top=0.5ex,bottom=0.5ex]
\fontsize{9.}{9.}\selectfont
\begin{tabular}{ccc}
        \includegraphics[width=0.3\textwidth]{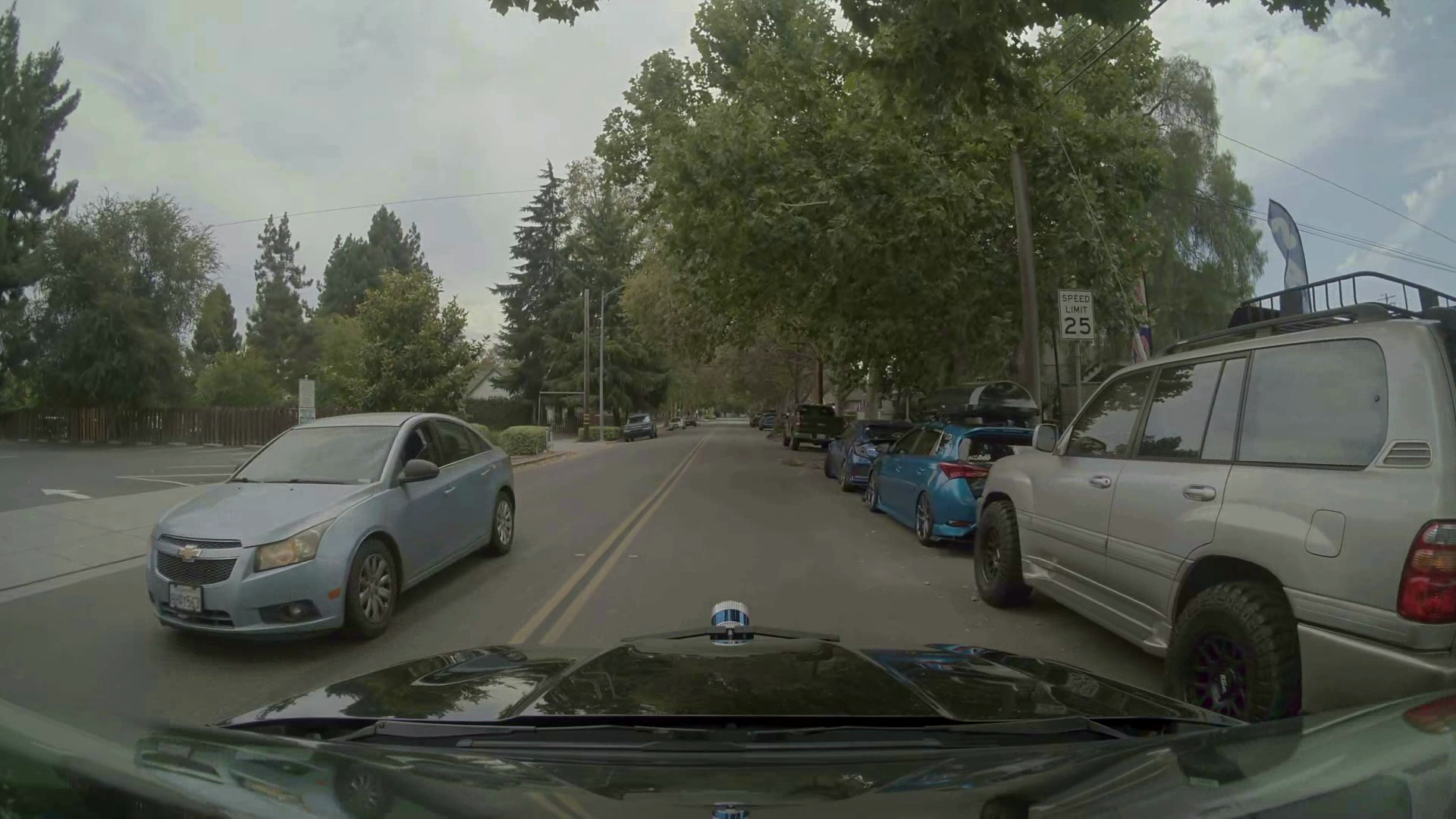} &
        \includegraphics[width=0.3\textwidth]{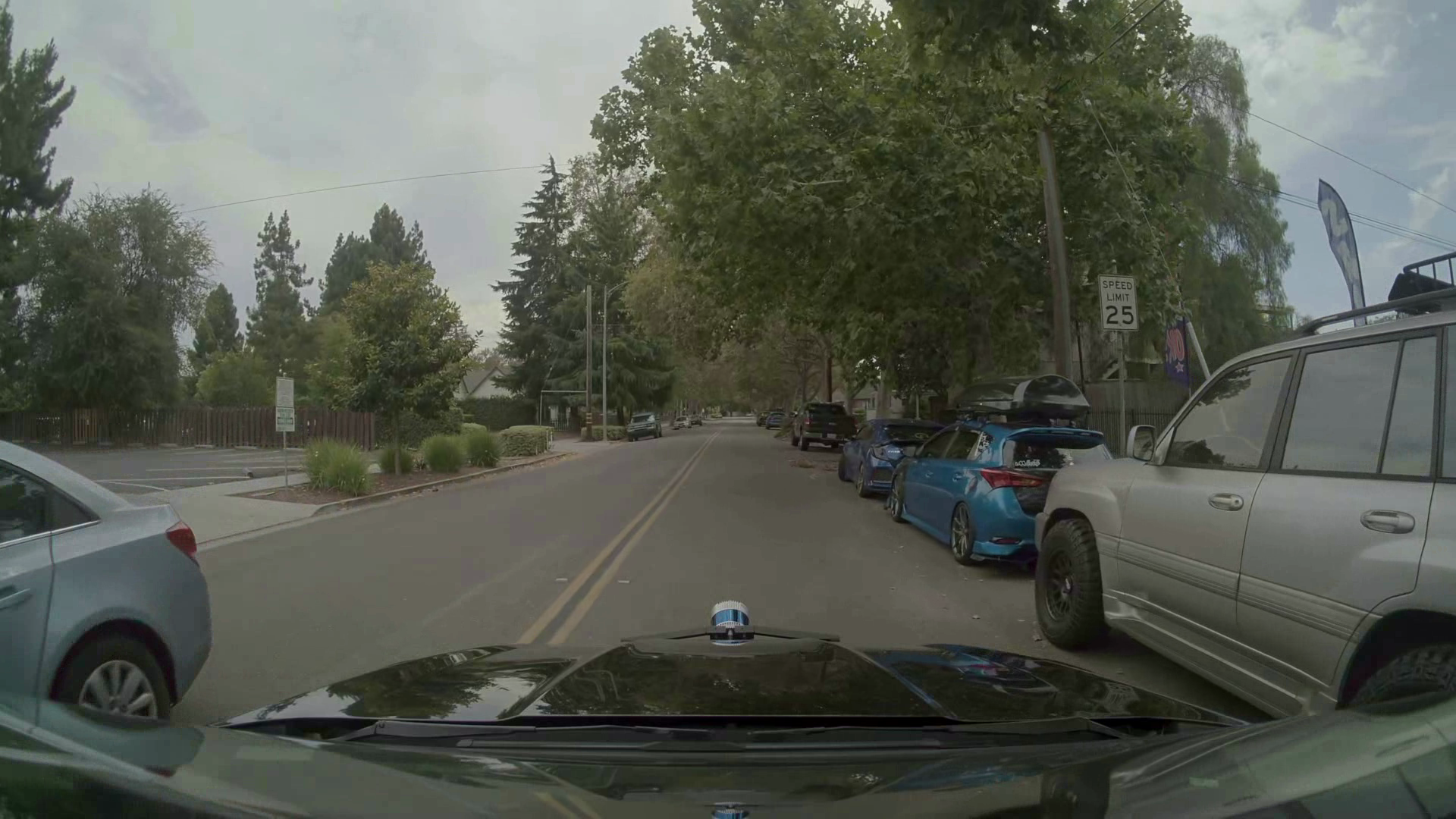} &
        \includegraphics[width=0.3\textwidth]{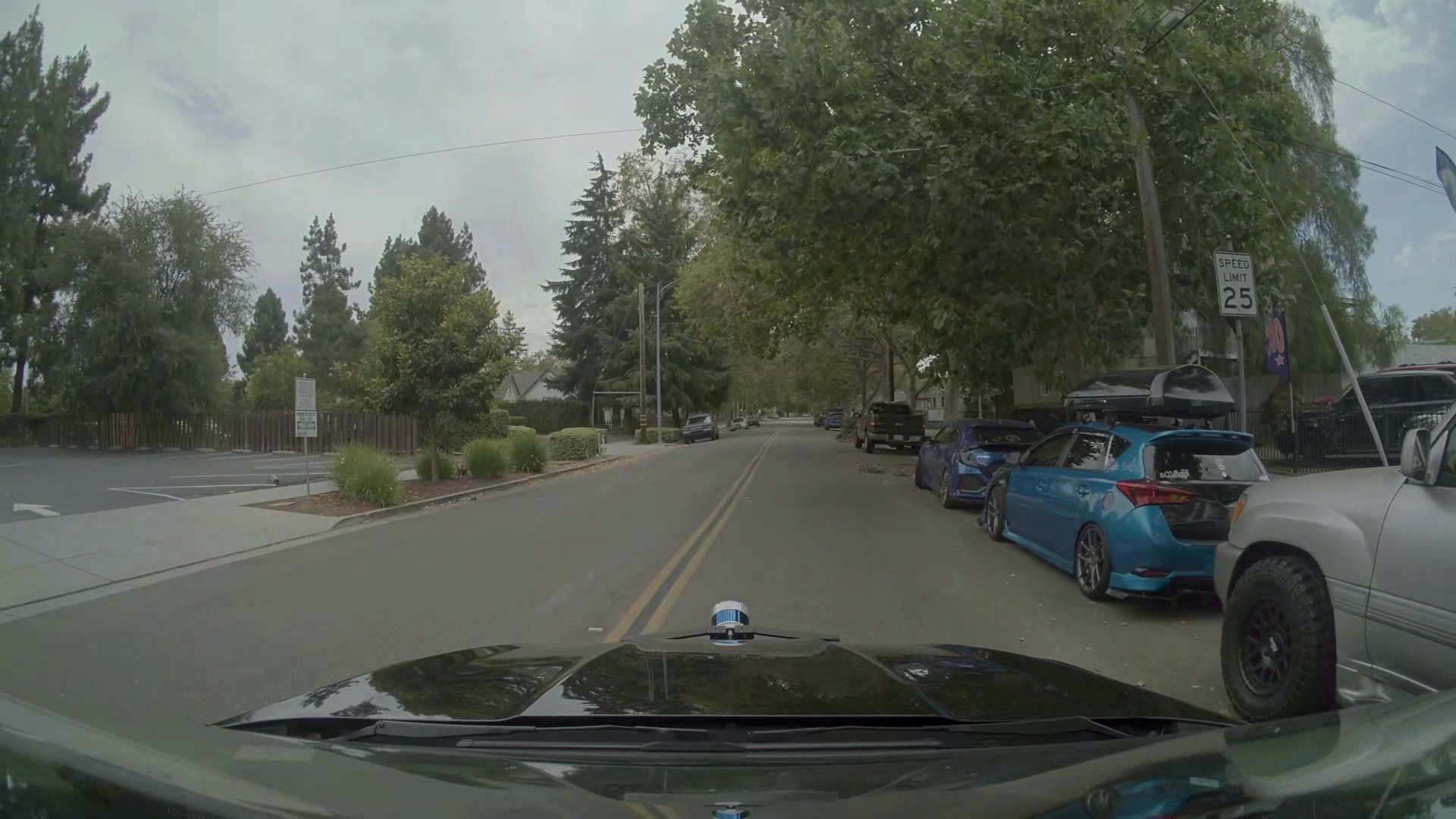} 
\end{tabular}
\begin{Verbatim}[breaklines=true, breaksymbolleft={}, breaksymbolright={}]
Based on the ego vehicle's actions in the video, what is the most likely immediate action it will take next?

A: turn right, B: turn left, C: change to right lane, D: change to left lane
\end{Verbatim}
\end{tcolorbox}

\begin{tcolorbox}[colback=cyan!3,colframe=nvidiagreen!75!white, title=\textsc{Response Before RL},left=0.5ex,right=0.5ex,top=0.5ex,bottom=0.5ex]
\fontsize{9.}{9.}\selectfont
\begin{Verbatim}[breaklines=true, breaksymbolleft={}, breaksymbolright={}]
Okay, let's break this down. The video shows the ego vehicle is driving on a single-lane road with a 25 mph speed limit. The road has parked cars on the right, so the driver is maintaining a safe distance there. The left side is a two-way divider with oncoming traffic, which is coming from the left in the video.

Since the road is single-lane, changing lanes (right or left) might not be possible. But the video shows a single-lane road, which usually means one lane in each direction, so there's no shoulder to the right for parking, but parked cars are already on the right. Maybe the road is a single lane each way, but parked cars are on the right side of that lane. So the driver is in the correct lane, just keeping a safe distance from the parked cars. The oncoming traffic is in the opposite lane, so the driver is staying in their lane. Changing to the left lane would be into oncoming traffic. So that's not possible. Therefore, the answer must be D.
\end{Verbatim}
\end{tcolorbox}

\begin{tcolorbox}[colback=blue!3,colframe=nvidiagreen!75!white, title=\textsc{Response After RL}, left=0.5ex,right=0.5ex,top=0.5ex,bottom=0.5ex]
\fontsize{9.}{9.}\selectfont
\begin{Verbatim}[breaklines=true, breaksymbolleft={}, breaksymbolright={}]
Okay, let's break this down. The user provided a driving scenario and asked for the most likely immediate action based on the ego vehicle's current actions. First, I need to recall the details given. The ego vehicle is on a narrow, two-way residential street with parked cars on both sides. The driver is maintaining a steady speed, staying in their lane, and being cautious because there's oncoming traffic in the opposite lane. The speed limit is 25 mph. The parked cars limit visibility, so they have to be extra careful. The road markings are double yellow lines, meaning no passing, so lane changes aren't allowed. The driver is keeping a consistent pace and staying vigilant. 

Now, the question is about predicting the next immediate action. The options are turning right, left, changing to the right lane, or left lane. Let's consider each option.

Option A: Turn right. The video shows a two-way street. The road is narrow with parked cars on both sides, so turning right might not be possible.

Option B: Turn left. Similar reasoning as A. Unless there's a left turn ahead, but the video doesn't indicate any upcoming intersections or turns.

Option C: Change to right lane. The current lane is the right one since it's a two-way street. Changing to the right lane again doesn't make sense because there's already parked cars on both sides. So this option is probably not correct.

Option D: Change to left lane. Wait, but in a two-way street with double yellow lines, there's no left lane to change into. The driver is in their respective lane (right for their direction), and the opposite lane is for oncoming traffic. So changing to the left lane would be into oncoming traffic, which is dangerous and not allowed. So this option doesn't fit either.

Wait, maybe the question is a trick. The video shows the road markings are double yellow lines, which means no passing, so changing lanes is prohibited. Therefore, the driver cannot change lanes, so options C and D are impossible. The only remaining options are turning right or left. But since the video doesn't show any turns, and the driver is maintaining their path, perhaps the answer is that none of the options are correct, but since we have to choose, I have no choice but to select none. Therefore, the answer is none.
\end{Verbatim}
\end{tcolorbox}

\caption{Intriguing results before and after RL. When presented with an ambiguous question, we observe that after RL, our model learns to reject all provided options based on its knowledge.}
\label{fig:resut_example_interesting}
\end{figure}

Interestingly, we found that through RL, our model learns to carefully evaluate the provided choices and reject them all if the question is ambiguous. As shown in \Cref{fig:resut_example_interesting}, the model assesses the feasibility of each choice and adopts a conservative action that is not within the choices when ambiguity arises.

\subsubsection{Intuitive Physics Results}
\label{sec::rl_intuitive_physics}

\begin{figure}[p]
\centering
    
\begin{tcolorbox}[colback=nvidiagreen!3,colframe=nvidiagreen!75!white,title=\textsc{Intuitive Physics: Arrow of Time},left=0.5ex,right=0.5ex,top=0.5ex,bottom=0.5ex]
\fontsize{9.}{9.}\selectfont
\begin{center}
\begin{tabular}{cccc}
        \includegraphics[width=0.21\textwidth]{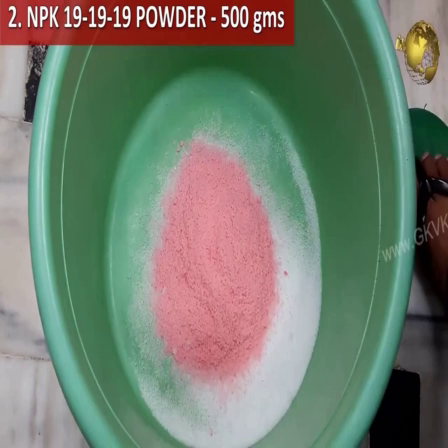} &
        \includegraphics[width=0.21\textwidth]{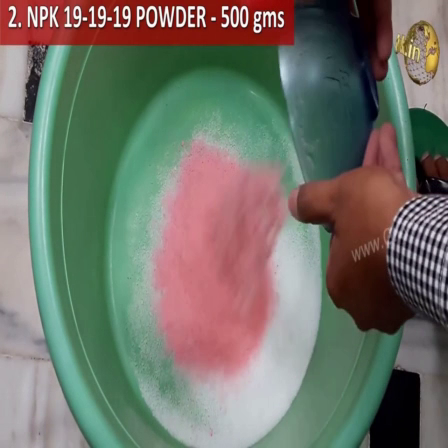} &
        \includegraphics[width=0.21\textwidth]{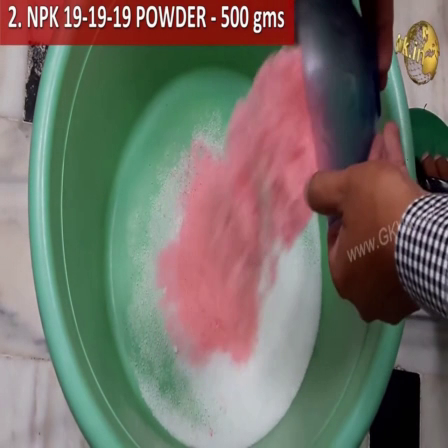} &
        \includegraphics[width=0.21\textwidth]{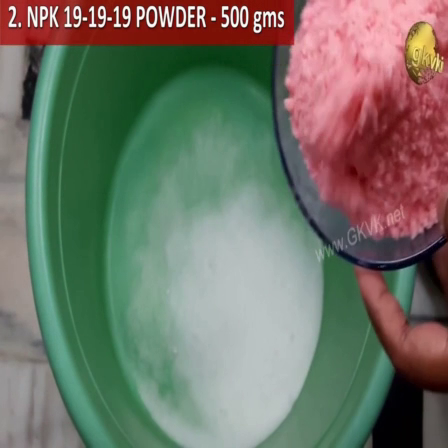} 
\end{tabular}
\end{center}
\begin{Verbatim}[breaklines=true, breaksymbolleft={}, breaksymbolright={}]
Detect whether the video plays forward or backward with confidence.

A: forward, B: backward
\end{Verbatim}
\end{tcolorbox}

\begin{tcolorbox}[colback=cyan!3,colframe=nvidiagreen!75!white, title=\textsc{Response Before RL},left=0.5ex,right=0.5ex,top=0.5ex,bottom=0.5ex]
\fontsize{9.}{9.}\selectfont
\begin{Verbatim}[breaklines=true, breaksymbolleft={}, breaksymbolright={}]
Okay, let's see. The user is asking whether the video plays forward or backward. The video shows a step-by-step process of mixing pink powder into a white liquid in a green bowl. The key points shown are the hand holding a blue container, pouring the pink powder into the bowl, and the camera being static.

Now, the question is about the direction of playback. Since the video shows a process where steps are performed in a logical sequence-like pouring the powder into the liquid-it's likely that the video is showing the correct order of events. If it were played backward, the powder would be disappearing from the bowl into the container, which doesn't make sense in this context. The actions shown, such as pouring and mixing, are typical of a forward process. The video does not show any reverse actions or objects moving in the opposite direction. The camera angle is static, so there's no indication of reverse motion effects. Therefore, the video should be playing forward.
\end{Verbatim}
\end{tcolorbox}

\begin{tcolorbox}[colback=blue!3,colframe=nvidiagreen!75!white, title=\textsc{Response After RL}, left=0.5ex,right=0.5ex,top=0.5ex,bottom=0.5ex]
\fontsize{9.}{9.}\selectfont
\begin{Verbatim}[breaklines=true, breaksymbolleft={}, breaksymbolright={}]
Okay, let's break this down. In the sequence of frames provided, the hand holding the container appears to be pouring a pink powder into a green bucket containing a white powder. 

As the sequence progresses, the amount of pink powder in the container increases while the amount in the bucket decreases. This suggests that the action is being undone, which is characteristic of a video playing in reverse. Additionally, the hand's motion and the position of the container change in a way that indicates the action is being retraced, further supporting the idea that the video is playing backward. 

The text on the screen remains constant throughout the frames, indicating that the content of the video is not changing, but rather the action being performed is being undone.

The text on the screen remains constant throughout the frames, indicating that the content of the video is not changing, but rather the action being performed is being undone.
\end{Verbatim}
\end{tcolorbox}

\caption{
While the model before RL struggles to understand and associate perception with reverse actions, RL enables the model to reason through time while avoiding distractions, such as stationary text.}
\label{fig:aot_example}
\end{figure}

Similar to self-supervised learning, a key advantage of intuitive physics is the ease of data scaling. 
Generating training data for the arrow of time requires only simple video reversal, spatial puzzles can be applied to any image to create challenging spatial continuity problems, and object permanence can be easily implemented in any simulated environment.
Moreover, these data sources can be seamlessly adapted for RL training with verifiable rewards.
With this recipe, we manage to create a larger RL dataset compared to common sense and embodied reasoning tasks with minimal human effort.

\Cref{tab:exp_rl_ssl} shows that with careful SFT data curation and targeted training, \cosmosreason-7B achieves significant improvements across all tasks, while Physical AI RL is able to further enhance spatial puzzle and object permanence capabilities. However, reasoning about the arrow of time still remains a challenge.

Through Physical AI RL, we found that the model further develops reasoning in space, time and object permanence.
\Cref{fig:aot_example} illustrates the differences in how Cosmos-Reason1 reasons about time before and after RL.
The model can identify anti-physics motions --- \emph{e.g.} powder defying gravity to rise into the bowl --- while remaining unaffected by stationary distractors in the video. This demonstrates that its reasoning extends beyond mere perception.
Similarly, in \Cref{fig:puzzle_example}, models without RL tend to conflate spatial questions with temporal reasoning. While they can perceive that the second frame lacks similarity to the first, their inherent biases lead them to default to video order, suggesting they rely more on temporal cues than genuine spatial understanding.
RL with spatial puzzles enables the model to extract key features from the first frame and systematically compare them across multiple frames, allowing it to accurately determine spatial relationships.
Lastly, \Cref{fig:object_example} shows that even with long CoT, the model without Phyiscal AI RL struggles with object permanence, often confusing itself when reasoning about an object's appearance and disappearance. In contrast, the RL model quickly reaches conclusions through direct and concise reasoning.

\section{Related Work}
\label{sec::related_work}

\subsection{Foundation Models for Physical AI}
Early approaches to integrating large pre-trained models into embodied agents primarily relied on using existing large language models (LLMs) and vision-language models (VLMs) as static modules. For example, the line of work using LLM as zero-shot task planner~\citep{li2022pre,song2023llmplanner,huang2022language,ahn2022can} extracts high-level action plans in natural language directly from pre-trained LLMs without additional fine-tuning, demonstrating the ability of LLMs to generalize across robotic tasks. Similarly, Code as Policies~\citep{codeaspolicies2022} extends the use of LLMs to generate structured code for robotic control. However, these pre-trained models, especially the VLMs, are not explicitly designed for physical interactions and, therefore, often lack the physical common sense understanding and result in suboptimal plans and execution. 

Another line of work focuses on directly training vision-language-action (VLA) models end-to-end for embodied agents~\citep{brohan2023rt,driess2023palme,kim2024openvla,yang2025magma,gemini_robotics}. They often start with a pre-trained vision-language model and fine-tune the model with robotics data to generate embodied actions. This approach has shown significant promise in improving generalization and robustness across robotic tasks. 
A notable subcategory of these methods introduces hierarchical policy representations that separate high-level language-driven planning from low-level action execution~\citep{shi2025hi,li2025hamster}. These system often use a high-level VLM interpret natural language instructions and generate plans, while low-level VLA execute fine-grained motor control. These hierarchical architectures improve task and decomposition. 

Recent efforts have sought to enable embodied AI with stronger reasoning abilities. One such approach is the embodied chain-of-thought (CoT) framework~\citep{zawalski2024ecot}, which enables robotic agents to reason through sequential decisions before executing actions. This aligns with broader trends in AI where explicit reasoning mechanisms improve interpretability and adaptability.
Similarly,~\citet{liu2023reflect,elhafsi2023semantic} introduce methods that perform step-by-step reasoning for explaining and correcting failures. While these methods advance the cognitive capabilities of embodied AI, most still rely on manual prompting to structure their reasoning processes, limiting their autonomous adaptation and generalization. CoT-VLA \citep{zhao2025cotvlavisualchainofthoughtreasoning} introduces a CoT process which first predicts a future image and then predicts the action which should interact with the environment to achieve the predicted future image.

Beyond robotics, VLA models have been applied to other physical embodiments, such as autonomous driving. For example, CoVLA~\citep{arai2024covla} introduces a large-scale vision-language-action dataset specifically for self-driving applications, facilitating research into multimodal decision-making in autonomous systems.

\subsection{Vision Language Models}
The community has made significant strides in building vision language models. Prominent families of models include Flamingo~\citep{alayrac2022flamingo}, LLaVA~\citep{liu2023visual}, InternVL~\citep{chen2024expanding}, QwenVL~\citep{bai2025qwen2}, NVLM~\citep{dai2024nvlm}, Llama-3.2-Vision~\citep{grattafiori2024llama}. 
These vision language models typically adopt one of two common architectures: the decoder-only architecture, as seen in models like LLaVA~\citep{liu2023visual} and InternVL~\citep{chen2024internvl}, which integrates image tokens within the LLM’s self-attention layers, and the cross-attention-based architecture, exemplified by Flamingo~\citep{alayrac2022flamingo} and Llama-3.2-Vision~\citep{grattafiori2024llama}, where image tokens are processed through the LLM’s cross-attention layers.
\citet{dai2024nvlm} compares both architectures in a state-of-the-art setting and finds that the decoder-only architecture exhibits stronger reasoning capabilities in college-level multidisciplinary knowledge and mathematical reasoning tasks within a visual context.
Building on this, we adopt the decoder-only architecture to develop the reasoning model for Physical AI.

\subsection{LLMs and VLMs with Reasoning Capabilities}
Early studies have shown that large language models~(LLMs) exhibit basic reasoning capabilities in mathematics~\citep{cobbe2021training}, coding~\citep{chen2021evaluating}, and general reasoning tasks. These capabilities can be further enhanced through chain-of-thought prompting \citep{wei2022chain}.
Recently, OpenAI o1~\citep{openaio1, jaech2024openai} demonstrated that LLMs' reasoning capabilities in coding and mathematics can be significantly enhanced through large-scale reinforcement learning.
Notably, the open-sourced DeepSeek-R1~\citep{deepseekai2025deepseekr1} has shared its training methodology with the community, providing valuable insights into building high-performance reasoning models.
However, existing studies primarily focus on reasoning tasks related to coding, mathematics, and STEM fields~\citep{liu2024acemath}, even within multimodal reasoning settings~\citep{qvq-72b-preview}.
Recently, there has been a surge of efforts aimed at integrating R1's reasoning capabilities into VLMs~\citep{liu2025visual, zhou2025r1, zhao2025r1, huang2025vision, wang2025v1}.
In this work, we explore reasoning capabilities in the context of Physical AI.

\section{Conclusion}
\label{sec::conclusion}

In this work, we present \cosmosreason, a family of multimodal large-language models specialized in physical world understanding and reasoning.
To specialize the model in Physical AI, we define ontologies to encapsulate foundational capabilities for Physical AI models and construct supervised fine-tuning data and benchmarks for common sense and embodied reasoning accordingly.
We further explore the idea of Physical AI RL crafting rule-based, verifiable rewards and using reinforcement learning to improve the model's capabilities with reasoning about space, time, and intuitive physics. Our experiments show that Physical AI SFT improves the backbone VLM's performance by more than 10\% on the proposed physical common sense and embodied reasoning benchmarks. Physical AI RL further boosts accuracy by over 5\%.
With Physical AI SFT and RL, \cosmosreason can learn intuitive physics, such as the arrow of time and object permanence, which existing models struggle with. We will make our code open-source and models open-weight to expedite the progress of building Physical AI systems that understand and perform complex tasks in the physical world.

\clearpage
\appendix
\section{Prompts Used for Data Curation}
\label{sec::prompts}

\subsection{Physical Common Sense Question-Construction Prompt}

\begin{tcolorbox}[colback=nvidiagreen!3,colframe=nvidiagreen!75!white,title=\textsc{Physical Common Sense Question-Construction Prompt},left=0.5ex,right=0.5ex,top=0.5ex,bottom=0.5ex]
\fontsize{9.}{9.}\selectfont
\begin{Verbatim}[breaklines=true, breaksymbolleft={}, breaksymbolright={}]
You will be given a detailed caption describing the video. Your task is to generate 6 extremely challenging questions to evaluate the reasoning ability of a state-of-the-art model that require multi-step deep reasoning from the caption.

Try your very best to use your creativity to generate extremely challenging questions!
Here is a list of categories of questions you should generate:
1. Common sense reasoning, including but not limited to:
- Physical common sense, such as gravity, balance, stability, support, elasticity, deformation, lighting, heat, motion, acceleration, etc.
- Physical attributes that are not directly mentioned in the caption, such as mass, temperature, etc.
- Object state changes, such as egg changed from liquid to solid, steak changed from raw to cooked, etc.
- Object permanence, such as object visibility, occlusion, etc.

2. Spatial reasoning, including but not limited to:
- Spatial plausibility, such as whether the object can be placed in a certain location, in a certain orientation, etc.
- Affordance, such as whether the object can be used for a certain purpose, etc.
- Scene or surrounding environment that is not directly mentioned in the caption, such as in a tunnel, underwater, weather (sunny, rainy, etc.), etc.

3. Temporal reasoning, including but not limited to:
- Complex action understanding, such as subtask or goal decomposition, whether a task is completed, etc.
- Temporal order of events, such as before/after/simultaneously, etc.
- Planning, such as whether the object can be used for a certain purpose, come up with a plan based on the video, what are the next steps, etc.

Below are some additional rules you must follow:
1. You must create questions that require both the information in the caption and the external knowledge to challenge the model's reasoning ability with your creativity.
2. You must NOT create any questions with answers that are directly given in the caption.
3. You must NOT create any questions that can be answered by external knowledge only without the information from the video caption.
4. When asking questions, you should give as little information as possible. The model you are evaluating on is expected to get any information needed to answer the question from the video itself.

In your 6 questions, 2 of them should be about common sense reasoning and planning using world knowledge, 2 of them should be about spatial reasoning, and 2 of them should be about temporal reasoning.

You question should be concise with a maximum of 10 words.

This is the caption:
{caption}

You should treat video caption as the video. Focus on the video itself and do not mention anything related to the captions. For example, you should not mention "the caption", "the description", "what is mentioned", etc.

Instead, you can use wordings like "in the video", "the video shows", etc.
\end{Verbatim}
\label{prompt:commonsense_reasoning_question_construction}
\end{tcolorbox}

\subsection{Physical Common Sense Reasoning Extraction Prompt}

\begin{tcolorbox}[colback=nvidiagreen!3,colframe=nvidiagreen!75!white,title=\textsc{Physical Common Sense Reasoning Extraction Prompt},left=0.5ex,right=0.5ex,top=0.5ex,bottom=0.5ex]
\fontsize{9.}{9.}\selectfont
\begin{Verbatim}[breaklines=true, breaksymbolleft={}, breaksymbolright={}]
This is the video you see:
This video showcases noodles being prepared. 

The video begins with someone standing by a large boiling pot and putting noodles into one of six small circular compartments dipped into boiling water in a large circular stainless steel pot. The compartments are also stainless steel. The person using their right hand, stirs the noodles in a circular motion using light brown wooden chopsticks, and their left holds the compartment into place using a dark metallic stick-like object. The pot is placed on a black burner. Steam rises from the pot, and bubbles form, indicating active boiling. The person cooking is wearing a grey top with the sleeves reaching his elbow and has a rope tied to their waist.

The setup includes a stainless steel surface stained with white drops, a silver pot filler tap above the boiling pot, and on its left there are two deep silver filters. A stainless steel pot with a stainless steel lid on the cooking pot's left, and a white cloth placed on the lid. The background also includes a big spoon next to the burner used to cook the noodles. The lighting is bright, causing reflections and shadows over the silver surface. The camera keeps alternating a push-in and a push-out motion in a medium shot over a high angle, providing a clear, unobstructed view of the cooking process.

Answer the following question:
What would happen if the pot was not boiling?
\end{Verbatim}
\label{prompt:commonsense_reasoning_extraction}
\end{tcolorbox}

\subsection{AgiBot Reasoning Extraction Prompt}

\begin{tcolorbox}[colback=nvidiagreen!3,colframe=nvidiagreen!75!white,title=\textsc{AgiBot Reasoning Extraction Prompt},left=0.5ex,right=0.5ex,top=0.5ex,bottom=0.5ex]
\fontsize{9.}{9.}\selectfont
\begin{Verbatim}[breaklines=true, breaksymbolleft={}, breaksymbolright={}]
The state of the environment is as follows:
Inside a supermarket near a fruit stand. Well-lit indoor environment with artificial lighting. The floor appears to be tiled, and the overall ambiance suggests a clean and organized retail space. 

Camera - Positioned at a medium height, slightly above the fruit stand, capturing the robot arms and the fruit stand from a frontal perspective. The camera provides a clear, steady shot focusing on the interaction between the robot arms and the fruits. It captures the details of the objects in the foreground while keeping the background slightly out of focus. The camera remains stationary, providing a stable view of the robot arms and the fruit stand. 

Fruit Stand - Center of the frame, consisting of wooden trays filled with various fruits. The fruit stand has multiple compartments made of light-colored wood. It contains a variety of fruits such as bananas, oranges, apples, pears, and others, arranged neatly in rows. 

Shopping Cart - In the foreground, partially visible under the robot arms. A standard metal shopping cart with red handles and a basket containing some items, including a plastic bag where the apple is being placed. 

Plastic Bag - Hanging from the shopping cart, below the robot arms. A transparent plastic bag with some red items inside, likely other fruits or vegetables. Hanging from the shopping cart, partially filled with items. Receives the red apple dropped by Robot Arm 2. 

Grapes - Held by Robot Arm 1, positioned over the fruit stand. A cluster of dark purple grapes, appearing fresh and ripe. 

Apple - Held by Robot Arm 2, positioned over the plastic bag in the shopping cart. A single red apple, shiny and smooth, indicating freshness.

A robot is performing the following task (may not be finished in the video): Pickup items in the supermarket

This is how the robot's actions impact the environment:

Robot Arm 1 - On the left side of the frame, extending towards the fruit stand. A mechanical arm with a black and white color scheme. It has a gripper mechanism at the end holding a bunch of grapes. The arm is articulated with joints allowing for precise movement. Holds a bunch of grapes with its grippers, positioned over the fruit stand.

Robot Arm 2 - On the right side of the frame, extending towards the fruit stand. Another mechanical arm similar in design to Robot Arm 1 but with a different configuration of the gripper mechanism. It is holding an apple and appears to be placing it into a plastic bag. Holds a red apple with its grippers, positioned above the plastic bag in the shopping cart. Releases the red apple into the plastic bag in the shopping cart.


**Output Requirements**:
- Predict the next immediate action of the robot.

**Response Format**:
- <action> your predicted action </action>.
\end{Verbatim}
\label{prompt:agibot_reasoning}
\end{tcolorbox}

\subsection{Spatial Puzzle Reasoning Extraction Prompt}

\begin{tcolorbox}[colback=nvidiagreen!3,colframe=nvidiagreen!75!white,title=\textsc{Spatial Puzzle Reasoning Extraction Prompt},left=0.5ex,right=0.5ex,top=0.5ex,bottom=0.5ex]
\fontsize{9.}{9.}\selectfont
\begin{Verbatim}[breaklines=true, breaksymbolleft={}, breaksymbolright={}]
You will be given a puzzle, which has 8 images, each broken into a 2x2 grid: top-left, top-right, bottom-left, bottom-right.
As a result, there are 32 frames and the first frame shows that "{first_frame_caption}".

{frames_and_captions}

For the remaining 31 frames, which three are most likely to be from the same images as the first frame? Then among the three, which one is most likely to be at {direction} of the first frame?

Give your answer in the following format:
<answer>
Same image: Frame a, Frame b, Frame c
{direction}: Frame d
</answer>
\end{Verbatim}
\label{prompt:puzzle_reasoning_prompt}
\end{tcolorbox}

\subsection{Arrow of Time Reasoning Extraction Prompt}

\begin{tcolorbox}[colback=nvidiagreen!3,colframe=nvidiagreen!75!white,title=\textsc{Arrow of Time Reasoning Extraction Prompt for forward video, backward video prompt is similar},left=0.5ex,right=0.5ex,top=0.5ex,bottom=0.5ex]
\fontsize{9.}{9.}\selectfont
\textbf{\underline{HUMAN TURN:}}
\begin{Verbatim}[breaklines=true, breaksymbolleft={}, breaksymbolright={}]
Here is a sequence of frames. Is it playing forward or backward?
\end{Verbatim}

\noindent\hdashrule{\linewidth}{0.4pt}{2pt}

\textbf{\underline{ASSISTANT TURN:}}
\begin{Verbatim}[breaklines=true, breaksymbolleft={}, breaksymbolright={}]
The video is playing forward.
\end{Verbatim}
\noindent\hdashrule{\linewidth}{0.4pt}{2pt}
\textbf{\underline{HUMAN TURN:}}
\begin{Verbatim}[breaklines=true, breaksymbolleft={}, breaksymbolright={}]
So here is a forward-playing video. Describe in details what visual clues do you notice that indicate the time is going forward i.e. the video is not reversed? Please avoid repeating the question and refrain from using bullet points. Present the evidence but no need to give the verdict.
\end{Verbatim}
\label{prompt:aot_reasoning_prompt}
\end{tcolorbox}

\subsection{Object Permanence Reasoning Extraction Prompt}

\begin{tcolorbox}[colback=nvidiagreen!3,colframe=nvidiagreen!75!white,title=\textsc{Object Permanence Reasoning Extraction Prompt},left=0.5ex,right=0.5ex,top=0.5ex,bottom=0.5ex]
\fontsize{9.}{9.}\selectfont
\begin{Verbatim}[breaklines=true, breaksymbolleft={}, breaksymbolright={}]
This is a video of a robotic simulation environment. The robotic arm moves and may occlude objects. The camera moves around the scene and then returns to a position near its initial location. Occlusion can also occur due to the camera's movement.

The objects in the videos are akita black bowl and akita black bowl and cookies and glazed rim porcelain ramekin and plate and wooden cabinet and flat stove.

akita black bowl is occluded in the middle of video and then vanishes unexpectedly.

Is there an object that becomes temporarily occluded but does not reappear in the end, contradicting object permanence?

Please answer the question in the following format: <think> your reasoning </think> <answer> your answer </answer>.
\end{Verbatim}
\label{prompt:permanence_reasoning_prompt}
\end{tcolorbox}

\newpage
\section{Additional results}
\label{sec::additional_results}

\begin{figure}[H]
\centering
\begin{tcolorbox}[colback=nvidiagreen!3,colframe=nvidiagreen!75!white,title=\textsc{Intuitive Physics: Spatial Puzzle},left=0.5ex,right=0.5ex,top=0.5ex,bottom=0.5ex]
\fontsize{7.}{7.}\selectfont
\begin{center}
\begin{tabular}{cccc}
    \textbf{Frame 1} & \textbf{Frame 2} & \textbf{Frame 9} & \textbf{Frame 11} \\
    \includegraphics[width=0.12\textwidth]{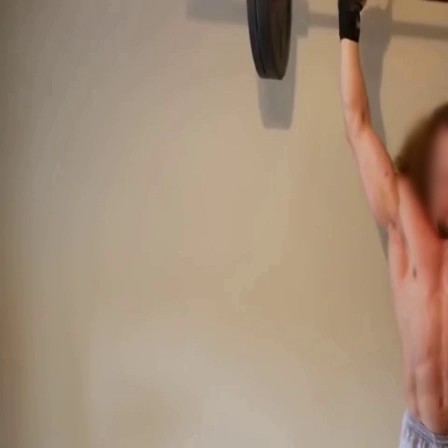} &
    \includegraphics[width=0.12\textwidth]{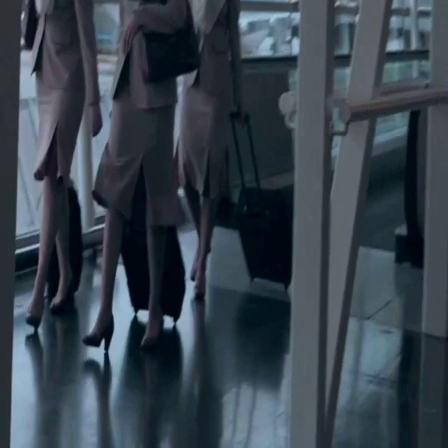} &
    \includegraphics[width=0.12\textwidth]{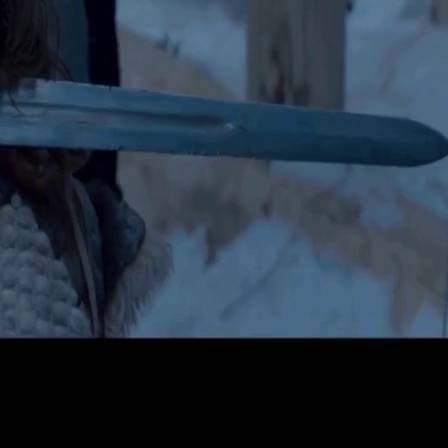} &
    \includegraphics[width=0.12\textwidth]{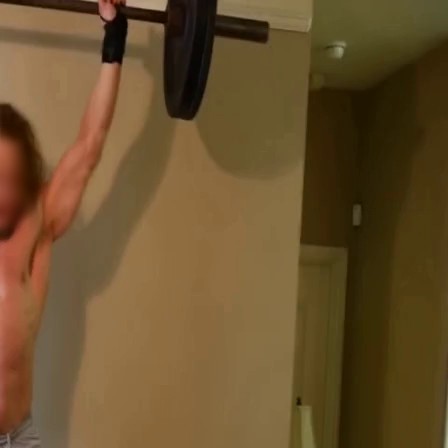} \\
    \textbf{Frame 14} & \textbf{Frame 17} & \textbf{Frame 24} & \textbf{Frame 30} \\
    \includegraphics[width=0.12\textwidth]{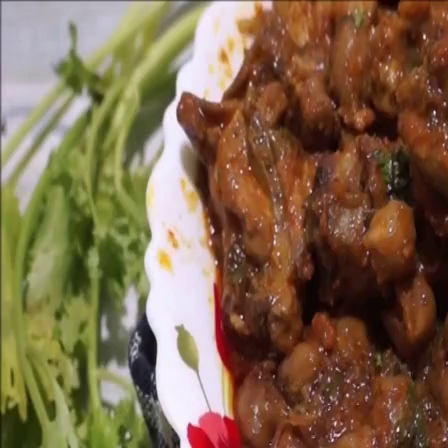} &
    \includegraphics[width=0.12\textwidth]{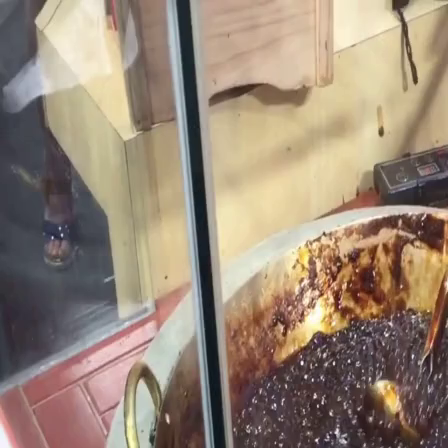} &
    \includegraphics[width=0.12\textwidth]{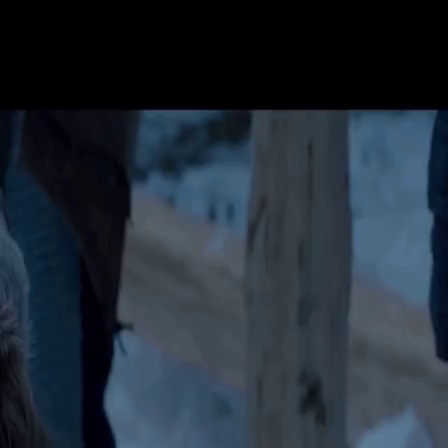} &
    \includegraphics[width=0.12\textwidth]{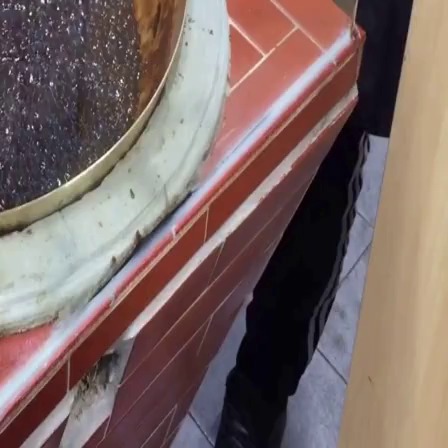} \\
\end{tabular}
\end{center}
\begin{Verbatim}[breaklines=true, breaksymbolleft={}, breaksymbolright={}]
You are given 32 frames from different contexts. Looking at the first frame, which other frame is most likely to be at right of the first frame?

A: Frame 2, B: Frame 30, C: Frame 9, D: Frame 11
\end{Verbatim}
\end{tcolorbox}

\begin{tcolorbox}[colback=cyan!3,colframe=nvidiagreen!75!white, title=\textsc{Response Before RL},left=0.5ex,right=0.5ex,top=0.5ex,bottom=0.5ex]
\fontsize{7.}{7.}\selectfont
\begin{Verbatim}[breaklines=true, breaksymbolleft={}, breaksymbolright={}]
Okay, let's tackle this question. The user is asking which frame is most likely to be on the right of the first frame based on the given video description. First, I need to recall the details of each frame. The first frame shown is a close-up of a shirtless man lifting weights with a barbell overhead. The question is about the spatial relationship between frames, specifically which frame would be positioned to the right of this first frame.

Looking at the options provided: A is Frame 2, B is Frame 30, C is Frame 9, D is Frame 11. To figure out the correct answer, I need to check the order of the frames as listed in the video description. The frames are numbered sequentially from 1 to 32, and each frame's content gives a clue about its position relative to others.

The video starts with Frame 1 (the weightlifter), then moves through various scenes. Let's scan through the order to see which frame comes after the first one. For example, after Frame 1, the next frame might be Frame 2, which could be the two women walking in an airport. But the question is about the right side of the first frame, not necessarily the next frame in the sequence. However, since the frames are shown in order, and the answer options are based on their numerical sequence, it's likely that the next frame (Frame 2) would be adjacent or in the same general area, especially if the video is structured as a continuous sequence with adjacent scenes. Looking at the given frames, Frame 2 is the two women in an airport. If the first frame is the weightlifter, and the next logical scene is the women walking, then Frame 2 would be to the right. Alternatively, if the sequence jumps around, but given the structure of the video (as per the video), the frames are presented in order. Therefore, the right adjacent frame would be the next one in sequence, which is Frame 2. Other options like Frame 30 or 9 or 11 don't immediately follow Frame 1 in the video order. 

So the most logical choice based on the sequence provided is Frame 2.
\end{Verbatim}
\end{tcolorbox}

\begin{tcolorbox}[colback=blue!3,colframe=nvidiagreen!75!white, title=\textsc{Response After RL}, left=0.5ex,right=0.5ex,top=0.5ex,bottom=0.5ex]
\fontsize{7.}{7.}\selectfont
\begin{Verbatim}[breaklines=true, breaksymbolleft={}, breaksymbolright={}]
Okay, let's break this down. To determine which frame is most likely to be to the right of the first frame, we need to identify frames that share similar characteristics with the first frame. The first frame depicts a shirtless man with long hair and a beard doing pull-ups in a minimalist room. Key elements include the man's appearance, the pull-up bar, and the room's simplicity.

We examine the other frames for matching details. Frame 11 shows a shirtless man with curly hair and a beard doing pull-ups in a room with beige walls and a closed door. This closely matches the first frame, particularly the man's appearance and the room's features. Both frames involve pull-ups and share a minimalist aesthetic.

Other frames do not align as closely. Frame 4 shows a child on a chair, which is unrelated. Frame 17 features a kitchen, which is a different setting. Frame 14 shows a person in a kitchen, also unrelated. Frame 24 depicts a snowy landscape, which is entirely different. Given the similarities between Frame 1 and Frame 11, it is reasonable to conclude that Frame 11 is the most likely candidate to be to the right of the first frame. Both frames depict the same man performing pull-ups in a similar setting, making them strong candidates for being part of the same image.

Therefore, the correct answer is  Frame 11.
\end{Verbatim}
\end{tcolorbox}

\caption{
While the model before RL mistakenly associates spatial questions with temporal reasoning, RL enables the model to identify key features of the first frame and compare them with subsequent frames.}
\label{fig:puzzle_example}
\end{figure}
\begin{figure}[p]
\centering
    
\begin{tcolorbox}[colback=nvidiagreen!3,colframe=nvidiagreen!75!white,title=\textsc{Intuitive Physics: Object Permanence},left=0.5ex,right=0.5ex,top=0.5ex,bottom=0.5ex]
\fontsize{7.}{7.}\selectfont
\begin{tabular}{cccccccc}
        \includegraphics[width=0.10\textwidth]{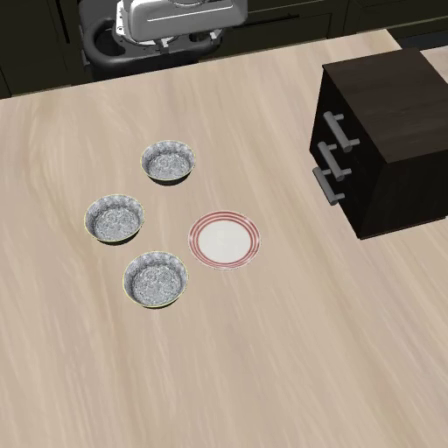} &
        \includegraphics[width=0.10\textwidth]{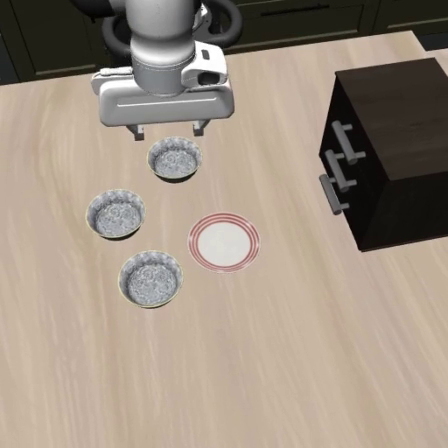} &
        \includegraphics[width=0.10\textwidth]{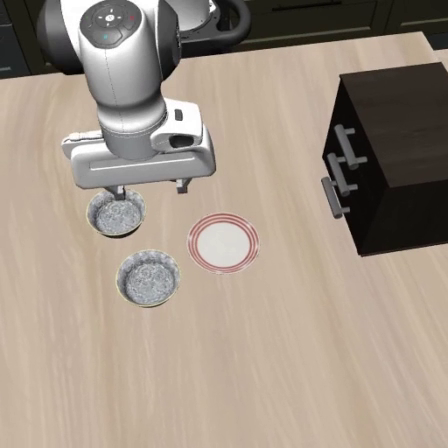} &
        \includegraphics[width=0.10\textwidth]{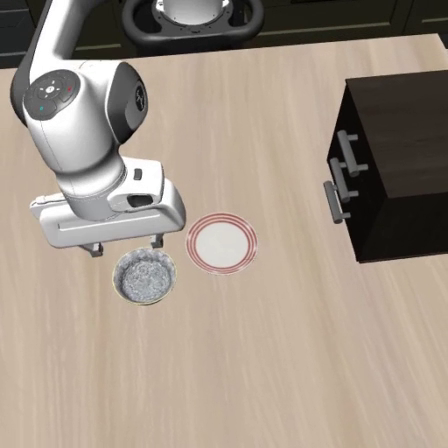} &
        \includegraphics[width=0.10\textwidth]{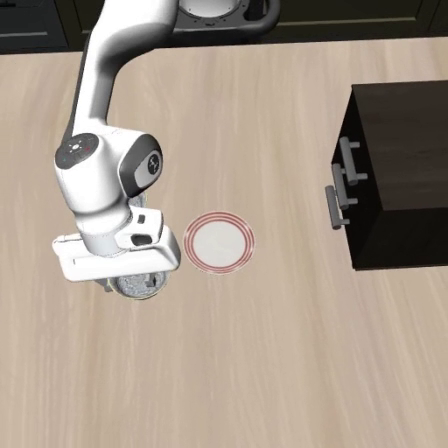} &
        \includegraphics[width=0.10\textwidth]{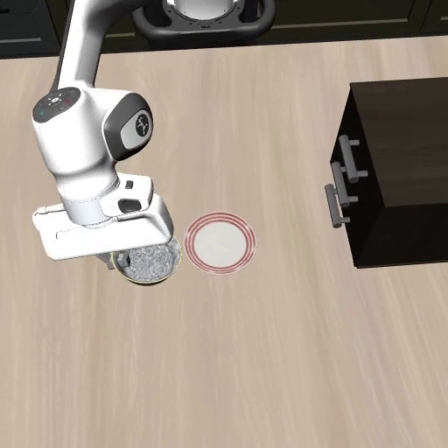} &
        \includegraphics[width=0.10\textwidth]{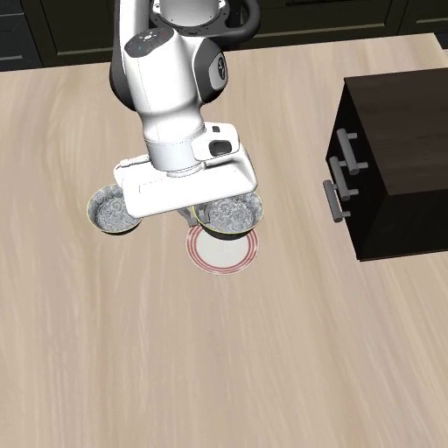} &
        \includegraphics[width=0.10\textwidth]{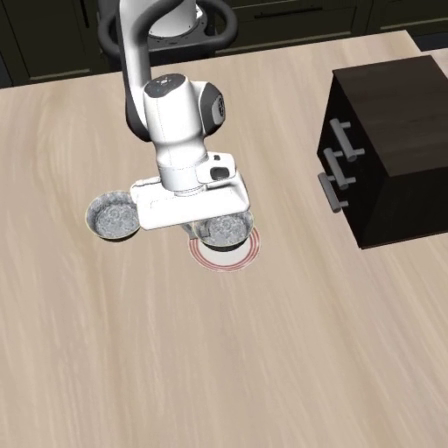} 
\end{tabular}
\begin{Verbatim}[breaklines=true, breaksymbolleft={}, breaksymbolright={}]
This is a video of a robotic simulation environment. The robotic arm moves and may occlude objects. The camera moves around the scene and then returns to a position near its initial location. Occlusion can also occur due to the camera's movement. Is there an object that becomes temporarily occluded but does not reappear in the end, contradicting object permanence?

A: Yes, B: No
\end{Verbatim}
\end{tcolorbox}

\begin{tcolorbox}[colback=cyan!3,colframe=nvidiagreen!75!white, title=\textsc{Response Before RL},left=0.5ex,right=0.5ex,top=0.5ex,bottom=0.5ex]
\fontsize{7.}{7.}\selectfont
\begin{Verbatim}[breaklines=true, breaksymbolleft={}, breaksymbolright={}]
Okay, let's break this down. The user is asking whether there's an object that gets temporarily hidden (occluded) but doesn't come back, which would challenge the idea of object permanence. First, I need to recall what the video shows. The setup has a robotic arm moving around three silver plates and a red plate. The camera moves around the scene and then goes back to its original spot. The key points here are about occlusion and object permanence. Object permanence is the concept that objects continue to exist even when they're not visible. So if something is occluded (temporarily hidden) but reappears later, that's consistent with object permanence. But if it's occluded and doesn't come back, that would contradict it.

Looking at the actions: the robotic arm picks up a silver plate and moves it towards the red plate. The camera moves around, possibly showing parts of the scene from different angles. The video shows that the camera returns to its original position. So when the arm moves the plate, the camera's movement might cause temporary occlusion (like when the arm is between the camera and the plate), but once the camera moves back, the plate should still be there. However, the question is about whether the object comes back after being occluded.

The video ends with the arm moving the plate towards the red plate. The camera returning to its initial position might mean that the arm's action of moving the plate is completed, but does the plate reappear? Since the arm is moving the plate, once it's moved, the plate is no longer in its original position. But object permanence isn't about the object's location changing; it's about whether the object is lost from view entirely and not found again. If the camera's movement causes temporary occlusion (like the arm blocking the plate from view), but when the camera moves back, the plate is still there (just not in view), then that's okay. The plate doesn't vanish and reappear; it's just not visible at a moment.

The key here is that the plate was moved by the arm. If during the camera's movement, the arm's action caused the plate to be occluded (like moving it out of view), but once the camera returns, the plate is no longer in the camera's frame but still exists elsewhere. However, since the camera returns to its original position, and the arm has moved the plate, the plate might not reappear in the camera's view again. But does that count as the object coming back? If the arm has moved it out of the scene entirely, then it's gone, which would contradict object permanence. But if it's just moved to another part of the table and the camera's original position doesn't show it again, but the plate is still there somewhere else, that's fine.

Wait, the question says "occluded but does not reappear in the end" So if the plate is moved by the arm and then the camera's movement doesn't show it again, but the plate is still in the video (like on the table but not in the camera's current view), then it's still there. So the plate didn't vanish, just not in view. So object permanence holds because the object still exists, even if not seen. Therefore, the answer would be "No", because even if occluded temporarily, the object remains present and would reappear if the camera or arm moves back. The plate is moved but still exists in the video, just not in the camera's current view. So temporary occlusion doesn't contradict object permanence. Therefore, the answer is B.
\end{Verbatim}
\end{tcolorbox}

\begin{tcolorbox}[colback=blue!3,colframe=nvidiagreen!75!white, title=\textsc{Response After RL}, left=0.5ex,right=0.5ex,top=0.5ex,bottom=0.5ex]
\fontsize{7.}{7.}\selectfont
\begin{Verbatim}[breaklines=true, breaksymbolleft={}, breaksymbolright={}]
Okay, let's break this down. The question is asking if there's an object that becomes temporarily occluded ( meaning hidden from view for a bit ) but then doesn't reappear, which would contradict the idea of object permanence. Object permanence is the concept that objects exist even when not seen, so if something disappears and doesn't come back, that's a problem.
Looking at the video: The akita black bowl is shown as being occluded in the middle of the video and then vanishes unexpectedly. The key points here are that it's occluded (so temporarily hidden) and then it disappears. The question is whether the disappearance contradicts object permanence.

Object permanence would mean that even if something is hidden, it should still exist. But in the video, the bowl vanishes unexpectedly. So if it's no longer there, that would mean it's gone, which would contradict object permanence. But wait, maybe the bowl was moved out of the camera's view, but the video shows vanishes unexpectedly, implying it's not just moving out of the camera but actually disappearing.

So the answer would be A, yes, because the bowl disappears and doesn't reappear, which would contradict object permanence. The other objects like the plates and cabinet are shown as remaining, so only the bowl is the one that disappears. Therefore, the correct answer is A.
\end{Verbatim}
\end{tcolorbox}

\caption{
While the model before RL fails to reason about object permanence through long CoT, RL enables the model to correctly infer that the object's disappearance is not due to camera movement, using concise and direct reasoning.}
\label{fig:object_example}
\end{figure}

\clearpage
\section{Contributors and Acknowledgments}
\label{sec::contributors}

\subsection{Core Contributors}
Alisson Azzolini, 
Junjie Bai,
Jiaxin Cao,
Prithvijit Chattopadhyay,
Huayu Chen, 
Yin Cui, 
Yifan Ding,
Liang Feng, 
Siddharth Gururani, 
Imad El Hanafi, 
Zekun Hao, 
Jacob Huffman, 
Jingyi Jin, 
George Kurian, 
Nayeon Lee, 
Zhaoshuo Li, 
Xuan Li, 
Maosheng Liao, 
Tsung-Yi Lin, 
Ming-Yu Liu, 
Xiangyu Lu, 
Wei Ping, 
David W. Romero,
Shuran Song, 
Lyne Tchapmi, 
Andrew Z. Wang, 
Boxin Wang, 
Haoxiang Wang, 
Fangyin Wei, 
Jiashu Xu, 
Dinghao Yang,
Xiaodong Yang, 
Zhuolin Yang, 
Jingxu Zhang, 
Xiaohui Zeng, 
Zhe Zhang

\noindent \textbf{Contributions:} 
\textbf{YC, SS, MYL, TYL} defined physical common sense and embodied reasoning ontologies. \textbf{YC, PC, JX, AZW, TYL} curated physical common sense data and benchmarks. \textbf{XY, PC, FW, XL, AZW, SG, TYL} curated embodied reasoning data and benchmark. \textbf{ZL, JJ, TYL} designed the captioning method. \textbf{JX, ZH, LT, JJ, XL, ZL, SG, TYL} curated self-supervised data and benchmark. \textbf{YC, PC, SG} post-processed extracted reasoning traces. \textbf{BW, NL, ZY, WP} trained the base VLM models. \textbf{DWR, HW, XZ} built the infrastructure for supervised fine-tuning. \textbf{XZ, GK, AA, HW, SG, ZL, DWR, FW, TYL} trained the Physical AI supervised fine-tuned models. \textbf{ZL, NL, FW, YD} evaluated the reasoning models. \textbf{JH, IH, ZZ, HW, ZH, DWR, AA, JC, LF, ML, XLu, DY, JZ, JB} built the reinforcement learning infrastructure. \textbf{ZH, JH, HC, JX, LT}  trained Physical AI reinforcement learning models.  \textbf{YC, TYL} organized paper writing. \textbf{MYL, TYL} designed the overall system.

\subsection{Contributors}
Hannah Brandon,
Jinju Chu,
Jenna Diamond,
Francesco Ferroni,
Rama Govindaraju,
Jinwei Gu,
Brendan Johnson,
Rizwan Khan,
Elena Lantz,
Yen-Chen Lin,
Alice Luo,
Andrew Mathau,
Yun Ni,
Lindsey Pavao,
Misha Smelyanskiy,
Yao Xu

\textbf{Contributions:}
\textbf{YX, LP, AM, RK, JC, BJ, EL, HB, JD} helped collect human annotations. \textbf{JG} helped with the vision encoder. \textbf{FF} curated training data. \textbf{YN, RG, MS} supported scaling model training infrastructure. \textbf{YCL} provided insights on physical common sense ontology. \textbf{AL} helped with overall data curation.

\subsection{Acknowledgments}
We'd like to thank Wenliang Dai, Guo Chen, Guilin Liu, Zhiding Yu, Mohammad Shoeybi, Andrew Tao, Bryan Catanzaro for discussion and data curation of general VLM training.
Xinshuo Weng, Boris Ivanovic for data curation of AV. Moo Jin Kim for setting up a simulation environment of object permanence data collection. Heng Wang for human annotation pipeline. Pooya Jannaty for AoT idea discussion.

\clearpage
\setcitestyle{numbers}
\bibliographystyle{plainnat}
\bibliography{main}

\end{document}